\documentclass[letterpaper,twocolumn,10pt]{article}
\usepackage{usenix}
\usepackage[utf8]{inputenc}
\usepackage[english]{babel}
\usepackage{graphicx}
\usepackage{bm}
\usepackage{dsfont}
\usepackage{microtype}
\usepackage{booktabs}
\usepackage{xcolor}
\usepackage{comment}
\usepackage{footnote}
\usepackage{subfigure}
\usepackage{caption}
\usepackage[yyyymmdd]{datetime} 

\usepackage{amsmath, amsfonts, amssymb, amsthm}
\usepackage{txfonts} 
\usepackage{verbatim}
\usepackage{url}
\usepackage{color}
\usepackage{algorithm2e}
\usepackage[noend]{algpseudocode}
\usepackage{multirow}

\usepackage{siunitx}
\newcommand{\cD}{\mathcal D}

\newcommand{\dataset}[1]{{\sf\small #1}}
\newcommand{\bfdataset}[1]{\small\textbf{\textsf{#1}}}

\ifx\BlackBox\undefined
\newcommand{\BlackBox}{\rule{1.5ex}{1.5ex}}  
\fi
\ifx\proof\undefined
\newenvironment{proof}{\par\noindent{\bf Proof\ }}{\hfill\BlackBox\\[2mm]}
\fi

\newcommand\shortsection[1]{\vspace*{6pt}{\noindent\bf #1.}}

\theoremstyle{definition}
\newtheorem{definition}{Definition}[section]

\makeatletter
\def\url@leostyle{%
  \@ifundefined{selectfont}{\def\UrlFont{\sf}}{\def\UrlFont{\small\sffamily}}}
\makeatother
\makeatletter
\def\url@beostyle{%
  \@ifundefined{selectfont}{\def\UrlFont{\sf}}{\def\UrlFont{\scriptsize\sffamily}}}
\makeatother

\renewcommand\shortsection[1]{\vspace{6pt}{\noindent\bf #1.}}

\newcommand{\ra}[1]{\renewcommand{\arraystretch}{#1}}
\newcommand{\subcap}[1]{\centering \footnotesize #1}

\newcommand{\cC}{\mathcal C}

\makesavenoteenv{table}
\makesavenoteenv{tabular}

\let\OLDthebibliography\thebibliography
\renewcommand\thebibliography[1]{
  \OLDthebibliography{#1}
  \setlength{\parskip}{5pt}
  \setlength{\itemsep}{0pt plus 0.3ex}
}

\hypersetup{
    colorlinks=false,
    linkcolor=black,
    filecolor=black,      
    urlcolor=black,
}

\theoremstyle{definition}

\title{
\vspace*{-0.5in}                    
{{\normalsize \rm In 28\textsuperscript{th} {\em USENIX Security Symposium}, Santa Clara, CA, August 2019 (Revised and Updated Version: \today) \hrule}}
 \vspace*{0.4in}
Evaluating Differentially Private Machine Learning in Practice}

\author{
{\rm Bargav\ Jayaraman and David\ Evans}\\
Department of Computer Science\\
University of Virginia\\
}

\begin{document}

\pagestyle{plain}
\pagenumbering{gobble}

\maketitle

\begin{abstract}
Differential privacy is a strong notion for privacy that can be used to prove formal guarantees, in terms of a privacy budget, $\epsilon$, about how much information is leaked by a mechanism. 
However, implementations of privacy-preserving machine learning often select large values of $\epsilon$ in order to get acceptable utility of the model, with little understanding of the impact of such choices on meaningful privacy. Moreover, in scenarios where iterative learning procedures are used, differential privacy variants that offer tighter analyses are used which appear to reduce the needed privacy budget but present poorly understood trade-offs between privacy and utility. In this paper, we quantify the impact of these choices on privacy in experiments with logistic regression and neural network models. Our main finding is that there is a huge gap between the upper bounds on privacy loss that can be guaranteed, even with advanced mechanisms, and the effective privacy loss that can be measured using current inference attacks. Current mechanisms for differentially private machine learning rarely offer acceptable utility-privacy trade-offs with guarantees for complex learning tasks: settings that provide limited accuracy loss provide meaningless privacy guarantees, and settings that provide strong privacy guarantees result in useless models.
\end{abstract}

\section{Introduction}\label{intro}

Differential privacy has become a de facto privacy standard, and nearly all works on privacy-preserving machine learning use some form of differential privacy. These works include designs for differentially private versions of prominent machine learning algorithms including empirical risk minimization~\cite{chaudhuri2009privacy, chaudhuri2011differentially} and deep neural networks~\cite{shokri2015privacy, abadi2016deep}. 

While many methods for achieving differential privacy have been proposed, it is not well understood how to use these methods in practice. In particular, there is little concrete guidance on how to choose an appropriate privacy budget $\epsilon$, and limited understanding of how variants of the differential privacy definition 
impact privacy in practice. As a result, privacy-preserving machine learning implementations tend to choose arbitrary values for $\epsilon$ as needed to achieve acceptable model utility. For instance, the implementation of Shokri and Shmatikov~\cite{shokri2015privacy} requires $\epsilon$ proportional to the size of the target deep learning model, which could be in the order of few millions. Setting $\epsilon$ to such arbitrarily large values severely undermines the value of the privacy guarantees provided by differential privacy, although there is no consensus on a hard threshold value for $\epsilon$ above which formal guarantees differential privacy provides become meaningless in practice.

One proposed way to improve utility for a given privacy budget is to tighten the composition of differential privacy. Several differential privacy variants have been proposed that can provide a tighter analysis of the privacy budget under composition~\cite{mironov2017renyi, bun2016concentrated, dwork2016concentrated}, so can achieve what appears to be more privacy (i.e., lower $\epsilon$ values) for the same amount of noise, thus providing better utility even for given $\epsilon$ values. 
How much privacy leaks in adversarial scenarios, however, is not well understood. We shed light on this question by evaluating the mechanisms using different differential privacy variants for different choices of $\epsilon$ values and empirically measuring the privacy leakage, including how many individual training records are exposed by membership inference attacks on different models.

\shortsection{Contributions} Our main contribution is the evaluation of differential privacy mechanisms for machine learning to understand the impact of different choices of $\epsilon$, across different variants of differential privacy, on both utility and privacy. We focus our evaluation on gradient perturbation mechanisms, which are applicable to a wide class of machine learning algorithms including empirical risk minimization (ERM) algorithms such as logistic regression and deep learning (Section \ref{sec:dp_ml}). Our experiments cover three popular differential privacy variants (summarized in Section \ref{dp_background}): differential privacy with advanced composition (AC), zero-concentrated differential privacy~\cite{bun2016concentrated} (zCDP), and R\'{e}nyi differential privacy~\cite{mironov2017renyi} (RDP). These variations allow for tighter analysis of cumulative privacy loss, thereby reducing the noise that must be added in the training process to satisfy a particular privacy budget. We evaluate the concrete privacy loss of these variations using membership inference attacks~\cite{shokri2017membership, yeom2018privacy} and attribute inference attacks~\cite{yeom2018privacy} (Section \ref{sec:attacks}). 

While the model utility increases with the privacy budget (for any otherwise fixed setting), increasing the privacy budget also increases the success rate of inference attacks. Hence, we aim to find values of $\epsilon$ that achieve a balance between utility and privacy, and to understand the impact of privacy leakage concretely. We study both logistic regression and neural network models, on two multi-class classification data sets. Our key findings (Section~\ref{sec:experiments}) quantify the practical risks of using different differential privacy notions across a range of privacy budgets. Our main result is that, at least for the settings we consider, there is a huge gulf between what can be guaranteed by any differential privacy mechanism and what can be observed by known attacks. For acceptable utility levels, the guarantees are essentially meaningless regardless of the differential privacy variant, although the observed leakage by attacks is still relatively low.

\shortsection{Related work} 
Orthogonal to our work, Ding et al.~\cite{ding2018detecting} and Hay et al.~\cite{hay2016principled} evaluate the existing differential privacy implementations for the \emph{correctness} of implementation. Whereas, we assume correct implementations and aim to evaluate the impact of the privacy budget and choice of differential privacy variant. While Carlini et al.~\cite{carlini2018secret} also explore the effectiveness of differential privacy against attacks, they do not explicitly answer what values of $\epsilon$ should be used nor do they evaluate the privacy leakage of the differential privacy variants. Li et al.~\cite{li2013membership} raise concerns about relaxing the differential privacy notion in order to achieve better overall utility, but do not evaluate the leakage. We perform a thorough evaluation of the differential privacy variations and quantify their leakage for different privacy budgets. The work of Rahman et al.~\cite{rahman2018membership} is most closely related to our work. It evaluates differential privacy implementations against membership inference attacks, but does not evaluate the privacy leakage of different variants of differential privacy. Recently, Liu et al.~\cite{liu2019investigating} reported on extensive hypothesis testing differentially private machine learning using the Neyman-Pearson criterion. They give guidance on setting the privacy budget based on assumptions about the adversary's knowledge considering different types of auxiliary information that an adversary can obtain to strengthen the membership inference attack such as prior probability distribution of data, record correlation, and temporal correlation. 

\section{Differential Privacy for Machine Learning} \label{background}
Next, we review the definition of differential privacy and several variants. Section~\ref{sec:dp_ml} surveys mechanisms for achieving differentially private machine learning. Section~\ref{sec:defense} summarizes applications of differential privacy to machine learning and surveys implementations' choices about privacy budgets.

\subsection{Background on Differential Privacy}\label{dp_background}

Differential privacy is a probabilistic privacy mechanism that provides an information-theoretic security guarantee. Dwork~\cite{dwork2008differential} gives the following definition:
\begin{definition}[$(\epsilon, \delta)$-Differential Privacy]
Given two neighboring data sets $D$ and $D'$ differing by one record, a mechanism $\mathcal{M}$ preserves $(\epsilon, \delta)$-differential privacy if
\[Pr[\mathcal{M}(D) \in S] \le Pr[\mathcal{M}(D') \in S] \times e^\epsilon + \delta\]
where $\epsilon$ is the privacy budget and $\delta$ is the failure probability.
\end{definition}
When $\delta = 0$ we achieve a strictly stronger notion of $\epsilon$-differential privacy. 

The quantity $$\ln \frac{Pr[\mathcal{M}(D) \in S]}{Pr[\mathcal{M}(D') \in S]}$$ is called the \emph{privacy loss}. 

One way to achieve $\epsilon$-DP and $(\epsilon, \delta)$-DP is to add noise sampled from Laplace and Gaussian distributions respectively, where the noise is proportional to the \emph{sensitivity} of the mechanism $\mathcal{M}$:
\begin{definition}[Sensitivity]
For two neighboring data sets $D$ and $D'$ differing by one record, the sensitivity of $\mathcal{M}$ is the maximum change in the output of $\mathcal{M}$ over all possible inputs:
\[\Delta \mathcal{M} = \max_{D, D', \|D - D'\|_1 = 1} \|\mathcal{M}(D) - \mathcal{M}(D')\|\]
\end{definition}
\noindent where $\|\cdot\|$ is a norm of the vector. Throughout this paper we assume $\ell_2$-sensitivity which considers the upper bound on the $\ell_2$-norm of $\mathcal{M}(D) - \mathcal{M}(D')$. 

\shortsection{Composition}
Differential privacy satisfies a simple composition property: when two mechanisms with privacy budgets $\epsilon_1$ and $\epsilon_2$ are performed on the same data, together they consume a privacy budget of $\epsilon_1 + \epsilon_2$. Thus, composing multiple differentially private mechanisms leads to a linear increase in the privacy budget (or corresponding increases in noise to maintain a fixed $\epsilon$ total privacy budget). 

\begin{table*}[tb]
\hyphenpenalty10000
    \centering 
    \small 
    \renewcommand*{\arraystretch}{1.6}
    \setlength{\tabcolsep}{8pt}
    \begin{tabular}{p{2.4cm}p{2.6cm}p{3.0cm}p{3.2cm}p{3.1cm}}
        \toprule \raggedright
        & \multicolumn{1}{c}{\textbf{\small Advanced Comp.}} 
        & \multicolumn{1}{c}{\textbf{\small Concentrated (CDP)}} & 
        \multicolumn{1}{c}{\textbf{\small Zero-Concentrated (zCDP)}} & 
        \multicolumn{1}{c}{\textbf{\small R\'{e}nyi (RDP)}} \\ \hline
        {\small Expected Loss} & \multicolumn{1}{c}{$\epsilon (e^\epsilon - 1)$} & \multicolumn{1}{c}{$\mu = \frac{\epsilon (e^\epsilon - 1)}{2}$} & \multicolumn{1}{c}{$\zeta + \rho = \frac{\epsilon^2}{2}$} & 
        \multicolumn{1}{c}{$2\epsilon^2$} \\ 
        {\small Variance of Loss} & \multicolumn{1}{c}{$\epsilon^2$} & \multicolumn{1}{c}{$\tau^2 = \epsilon^2$} & 
        \multicolumn{1}{c}{$2\rho = \epsilon^2$} & \multicolumn{1}{c}{$\epsilon^2$} \\ \hline
        {\small Convert from $\epsilon$\nobreakdash-DP} & \multicolumn{1}{c}{-} & 
        \multicolumn{1}{c}{$(\frac{\epsilon (e^\epsilon - 1)}{2}, \epsilon)$-CDP} & \multicolumn{1}{c}{$(\frac{\epsilon^2}{2})$-zCDP} & 
        \multicolumn{1}{c}{$(\alpha, \epsilon)$-RDP} \\
        {\small Convert to DP} & 
        \multicolumn{1}{c}{-} & 
        \multicolumn{1}{c}{$(\mu + \tau\sqrt{2\log(1/\delta)},  \delta)$\nobreakdash-DP$^{\dagger}$} & 
        \multicolumn{1}{c}{$(\zeta + \rho + 2\sqrt{\rho\log(1/\delta)}, \delta)$\nobreakdash-DP} & 
        \multicolumn{1}{c}{$(\epsilon + \frac{\log(1/\delta)}{\alpha - 1}, \delta)$\nobreakdash-DP} \\ \hline
        {\raggedright \small Composition of $k$ \\ $\epsilon$\nobreakdash-DP Mechanisms} & 
        {\centering $(\epsilon\sqrt{2k\log(1/\delta)}\;\;\;\;\;$ \\ $\;\;\; +\;k \epsilon (e^\epsilon - 1), \delta)$\nobreakdash-DP} & 
        {\centering $(\epsilon\sqrt{2k\log(1/\delta)}\;\;\;$ \\ $\;\;\;\;\;\; +\; k \epsilon (e^\epsilon - 1) / 2, \delta)$\nobreakdash-DP} & 
        {\centering $(\epsilon\sqrt{2k \log(1/\delta)} $\\ $\;\;\;\;\;\;\;\;\;\;\;\; +\; k\epsilon^2/2, \delta)$\nobreakdash-DP} & \multicolumn{1}{c}
        {$(4\epsilon\sqrt{2k\log(1/\delta)}, \delta)$-DP$^{\ddagger}$} \\ \hline
    \end{tabular}
        \caption{Comparison of Different Variations of Differential Privacy}
    \subcap{Advanced composition is an implicit property of DP and hence there is no conversion to and from DP.\\ $\dagger$. Derived indirectly via zCDP. $\ddagger$. When $\log(1 / \delta) \ge \epsilon^2 k$.}
    
    \label{tab:dp_variations_comparison}
\end{table*}

\shortsection{Variants} \label{relaxed_definitions}
Dwork~\cite{dwork2014algorithmic} showed that this linear composition bound on $\epsilon$ can be reduced at the cost of \emph{slightly} increasing the failure probability $\delta$. In essence, this relaxation considers the linear composition of \emph{expected} privacy loss of mechanisms which can be converted to a cumulative privacy budget $\epsilon$ with high probability bound. Dwork defines this as the \emph{advanced composition theorem}, and proves that it applies to any differentially private mechanism. 

Three commonly-used subsequent variants of differential privacy which provide improved composition are Concentrated Differential Privacy~\cite{dwork2016concentrated}, Zero Concentrated Differential Privacy~\cite{bun2016concentrated}, and R\'{e}nyi Differential Privacy~\cite{mironov2017renyi}. All of these achieve tighter analysis of cumulative privacy loss by taking advantage of the fact that the privacy loss random variable is strictly centered around an \emph{expected} privacy loss. The cumulative privacy budget obtained from these analyses bounds the worst case privacy loss of the composition of mechanisms with all but $\delta$ failure probability. This reduces the noise required to satisfy a given privacy budget and hence improves utility over multiple compositions. However, it is important to consider the actual impact the noise reductions enabled by these variants have on the privacy leakage, which is a main focus of this paper. 

Note that the variants are just different techniques for analyzing the composition of the mechanism---by themselves they do not impact the noise added, which is all the actual privacy depends on. What they do is enable a tighter analysis of the guaranteed privacy. This means for a fixed privacy budget, the relaxed definitions can be satisfied by adding less noise than would be required for looser analyses, hence they result in less privacy for the same $\epsilon$ level.  
Throughout the paper, for simplicity, we refer to the differential privacy variants but implicitly mean the mechanism used to satisfy that definition of differential privacy. Thus, when we say RDP has a given privacy leakage, it means the corresponding gradient perturbation mechanism has that privacy leakage when it is analyzed using RDP to bound its cumulative privacy loss.

Dwork et al.~\cite{dwork2016concentrated} note that the privacy loss of a differentially private mechanism follows a sub-Gaussian distribution. In other words, the privacy loss is strictly distributed around the expected privacy loss and the spread is controlled by the variance of the sub-Gaussian distribution. Multiple compositions of differentially private mechanisms thus result in the aggregation of corresponding mean and variance values of the individual sub-Gaussian distributions. This can be converted to a cumulative privacy budget similar to the advanced composition theorem, which in turn reduces the noise that must be added to the individual mechanisms. The authors call this \emph{concentrated differential privacy} (CDP)~\cite{dwork2016concentrated}:
\begin{definition}[Concentrated Differential Privacy]
A randomized algorithm $\mathcal{M}$ is $(\mu, \tau)$-concentrated differentially private if, for all pairs of adjacent data sets $D$ and $D'$,  
\[\cD_{\textit{subG}} (\mathcal{M}(D)\;||\;\mathcal{M}(D')) \le (\mu, \tau)\]
\end{definition}\noindent
where the sub-Gaussian divergence, $\cD_{\textit{subG}}$, is defined such that the expected privacy loss is bounded by $\mu$ and after subtracting $\mu$, the resulting centered sub-Gaussian distribution has standard deviation $\tau$. Any $\epsilon$-DP algorithm satisfies  $(\epsilon \cdot (e^\epsilon - 1)/2, \epsilon)$-CDP, however the converse is not true.

A variation on CDP, \emph{zero-concentrated differential privacy} (zCDP)~\cite{bun2016concentrated} uses R\'{e}nyi divergence as a different method to show that the privacy loss random variable follows a sub-Gaussian distribution: 

\begin{definition}[Zero-Concentrated Differential Privacy]
A randomized mechanism $\mathcal{M}$ is $(\xi, \rho)$-zero-concentrated differentially private if, for all neighbouring data sets $D$ and $D'$ and all $\alpha \in (1,\infty)$,
\[\cD_\alpha (\mathcal{M}(D)\;||\;\mathcal{M}(D')) \le \xi + \rho \alpha\]
where $\cD_\alpha (\mathcal{M} (D)\;||\;\mathcal{M} (D'))$ is the $\alpha$-R\'{e}nyi divergence between the distribution of $\mathcal{M}(D)$ and the distribution of $\mathcal{M}(D')$.
\end{definition}
$\cD_\alpha$ also gives the $\alpha$-th moment of the privacy loss random variable. For example, $\cD_1$ gives the first order moment which is the mean or the expected privacy loss, and $\cD_2$ gives the second order moment or the variance of privacy loss. There is a direct relation between DP and zCDP. If $\mathcal{M}$ satisfies $\epsilon$-DP, then it also satisfies $(\frac{1}{2}\epsilon^2)$-zCDP. Furthermore, if $\mathcal{M}$ provides $\rho$-zCDP, it is $(\rho + 2\sqrt{\rho \log(1/\delta)}, \delta)$-DP for any $\delta > 0$. 
 
The R\'{e}nyi divergence allows zCDP to be mapped back to DP, which is not the case for CDP. However, Bun and Steinke~\cite{bun2016concentrated} give a relationship between CDP and zCDP, which allows an indirect mapping from CDP to DP (Table \ref{tab:dp_variations_comparison}). 

The use of R\'{e}nyi divergence as a metric to bound the privacy loss leads to the formulation of a more generic notion of R\'{e}nyi differential privacy (RDP) that is applicable to any individual moment of privacy loss random variable:
\begin{definition}[R\'{e}nyi Differential Privacy~\cite{mironov2017renyi}]
A randomized mechanism $\mathcal{M}$ is said to have $\epsilon$-R\'{e}nyi differential privacy of order $\alpha$ (which can be abbreviated as $(\alpha, \epsilon)$-RDP), if for any adjacent data sets $D$, $D'$ it holds that
\[\cD_\alpha (\mathcal{M}(D)\;||\;\mathcal{M}(D')) \le \epsilon.\]
\end{definition}
The main difference is that CDP and zCDP linearly bound \emph{all} positive moments of privacy loss, whereas RDP bounds one moment at a time, which allows for a more accurate numerical analysis of privacy loss. 
If $\mathcal{M}$ is an $(\alpha, \epsilon)$-RDP mechanism, it also satisfies $(\epsilon + \frac{\log 1/\delta}{\alpha - 1}, \delta)$-DP for any $0 < \delta < 1$.

Table \ref{tab:dp_variations_comparison} compares the relaxed variations of differential privacy. For all the variations, the privacy budget grows sub-linearly with the number of compositions $k$. 

\shortsection{Moments Accountant}
Motivated by the variants of differential privacy, Abadi et al.~\cite{abadi2016deep} propose the \emph{moments accountant} (MA) mechanism for bounding the cumulative privacy loss of differentially private algorithms. The moments accountant keeps track of a bound on the moments of the privacy loss random variable during composition. Though the authors do not formalize this as a differential privacy variant, their definition of the moments bound is analogous to the R\'{e}nyi divergence~\cite{mironov2017renyi}. Thus, the moments accountant can be considered as an instantiation of R\'{e}nyi differential privacy. The moments accountant is widely used for differentially private deep learning due to its practical implementation in the TensorFlow Privacy library~\cite{tf-privacy} (see Section~\ref{sec:defense} and Table~\ref{tab:relaxednotions}). 

\subsection{Differential Privacy Methods for ML}\label{sec:dp_ml}

This section summarizes methods for modifying machine learning algorithms to satisfy differential privacy. First, we review convex optimization problems, such as empirical risk minimization (ERM) algorithms, and show several methods for achieving differential privacy during the learning process. Next, we discuss methods that can be applied to non-convex optimization problems, including deep learning.

\shortsection{ERM}
Given a training data set $(X, y)$, where $X$ is a feature matrix and $y$ is the vector of class labels, an ERM algorithm aims to reduce the convex objective function of the form,
\begin{equation*}
    J(\theta) = \frac{1}{n}\sum_{i=1}^n \ell(\theta, X_i, y_i) + \lambda R(\theta),
\end{equation*}
where $\ell(\cdot)$ is a convex loss function (such as mean square error (MSE) or cross-entropy loss) that measures the training loss for a given $\theta$, and 
$R(\cdot)$ is a regularization function. Commonly used regularization functions include $\ell_1$ penalty, which makes the vector $\theta$ sparse, and $\ell_2$ penalty, which shrinks the values of $\theta$ vector. 

The goal of the algorithm is to find the optimal $\theta^*$ that minimizes the objective function: $\theta^* = \arg\min_\theta J(\theta)$. While many first order~\cite{duchi2011adaptive, zeiler2012adadelta, kingma2014adam, polyak1992acceleration} and second order~\cite{liu1989limited, li2001modified} methods exist to solve this minimization problem, the most basic procedure is gradient descent where we iteratively calculate the gradient of $J(\theta)$ with respect to $\theta$ and update $\theta$ with the gradient information. This process is repeated until $J(\theta) \approx 0$ or some other termination condition is met. 

\begin{algorithm}[bt]
 \KwData{Training data set $(X, y)$}
 \KwResult{Model parameters $\theta$}
 $\theta \leftarrow \text{Init}(0)$\\
 \textcolor{red}{\#1. Add noise here: \emph{objective perturbation}}\\
 $J(\theta) = \frac{1}{n}\sum_{i=1}^n \ell(\theta, X_i, y_i) + \lambda R(\theta) \textcolor{red}{+ \beta}$\\
 \For{epoch {\bf{in}} epochs}{
  \textcolor{blue}{\#2. Add noise here: \emph{gradient perturbation}}\\
  $\theta = \theta - \eta \textcolor{blue}{(}\nabla J(\theta) \textcolor{blue}{+ \beta)}$\\
 }
 \textcolor{purple}{\#3. Add noise here: \emph{output perturbation}}\\
 \Return $\theta \textcolor{purple}{+ \beta}$\\[2ex]
 \caption{Privacy noise mechanisms.}
 \label{alg:dp_mechanisms}
\end{algorithm}

There are three obvious candidates for where to add privacy-preserving noise during this training process, demarcated in Algorithm~\ref{alg:dp_mechanisms}. First, we could add noise to the objective function $J(\theta)$, which gives us the \emph{objective perturbation mechanism} (\#1 in Algorithm \ref{alg:dp_mechanisms}). Second, we could add noise to the gradients at each iteration, which gives us the \emph{gradient perturbation mechanism} (\#2). Finally, we can add noise to $\theta^*$ obtained after the training, which gives us the \emph{output perturbation mechanism} (\#3). While there are other methods of achieving differential privacy such as input perturbation~\cite{duchi2013local}, sample-aggregate framework~\cite{nissim2007smooth}, exponential mechanism~\cite{mcsherry2007mechanism} and teacher ensemble framework~\cite{papernot2017semi}. We focus our experimental analysis on gradient perturbation since it is applicable to all machine learning algorithms in general and is widely used for deep learning with differential privacy.

The amount of noise that must be added depends on the sensitivity of the machine learning algorithm. For instance, consider logistic regression with $\ell_2$ regularization penalty. The objective function is of the form:
\begin{equation*}
    J(\theta) = \frac{1}{n}\sum_{i=1}^n \log(1 + e^{-X_i^\top \theta y_i}) + \frac{\lambda}{2}\, \|\;\theta\;\|_2^2
\end{equation*}
Assume that the training features are bounded, $\|X_i\|_2 \le 1$ and $y_i \in \{-1, 1\}$. Chaudhuri et al.~\cite{chaudhuri2011differentially} prove that for this setting, objective perturbation requires sampling noise in the scale of $\frac{2}{n\epsilon}$, and output perturbation requires sampling noise in the scale of $\frac{2}{n\lambda\epsilon}$. The gradient of the objective function is:
\begin{equation*}
    \nabla J(\theta) = \frac{1}{n}\sum_{i=1}^n \frac{-X_iy_i}{1 + e^{X_i^\top \theta y_i}} + \lambda \theta
\end{equation*}
which has a sensitivity of $\frac{2}{n}$. Thus, gradient perturbation requires sampling noise in the scale of $\frac{2}{n\epsilon}$ at each iteration.

\shortsection{Deep learning}
Deep learning follows the same learning procedure as in Algorithm~\ref{alg:dp_mechanisms}, but the objective function is non-convex. As a result, the sensitivity analysis methods of Chaudhuri et al.~\cite{chaudhuri2011differentially} do not hold as they require a strong convexity assumption. Hence, their output and objective perturbation methods are not applicable. An alternative approach is to replace the non-convex function with a convex polynomial function~\cite{phan2016differential, phan2017preserving}, and then use the standard objective perturbation. This approach requires carefully designing convex polynomial functions that can approximate the non-convexity, which can still limit the model's learning capacity. Moreover, it would require a considerable change in the existing machine learning infrastructure. 

A simpler and more popular approach is to add noise to the gradients. Application of gradient perturbation requires a bound on the gradient norm. Since the gradient norm can be unbounded in deep learning, gradient perturbation can be used after manually clipping the gradients at each iteration. As noted by Abadi et al.~\cite{abadi2016deep}, norm clipping provides a sensitivity bound on the gradients which is required for generating noise in gradient perturbation.

\subsection{Implementing Differential Privacy}\label{sec:defense}

\begin{table*}[tb]
    \centering
    \ra{1.1}
    \begin{tabular}{@{}cccS[table-format=6.0]S[table-format=3.0]S[table-format=2.3]@{}}
        \toprule
         & Perturbation & Data Set & $n$ & 
         \multicolumn{1}{c}{$d$} & \multicolumn{1}{c}{$\epsilon$} \\ \hline
        \multirow{2}{*}{Chaudhuri et al.~\cite{chaudhuri2011differentially}} & \multirow{2}{*}{Output and Objective} & Adult & 45220 & 105 & 0.2 \\
         & & KDDCup99 & 70000 & 119 & 0.2 \\ \hline
        Pathak et al.~\cite{pathak2010multiparty} & Output & Adult & 45220 & 105 & 0.2 \\ \hline
        \multirow{2}{*}{Hamm et al.~\cite{hamm2016learning}} & \multirow{2}{*}{Output} & KDDCup99 & 493000 & 123 & 1.0 \\ 
         & & URL & 200000 & 50 & 1.0 \\ \hline
        \multirow{2}{*}{Zhang et al.~\cite{zhang2012functional}} & \multirow{2}{*}{Objective} & US & 370000 & 14 & 0.8 \\ 
         & & Brazil & 190000 & 14 & 0.8 \\ \hline
        \multirow{2}{*}{Jain and Thakurta~\cite{jain2013differentially}} & \multirow{2}{*}{Objective} & CoverType & 500000 & 54 & 0.5 \\ 
         & & KDDCup2010 & 20000 & \multicolumn{1}{r}{2M} & 0.5 \\ \hline
        \multirow{2}{*}{Jain and Thakurta~\cite{jain2014near}} & \multirow{2}{*}{Output and Objective} & URL & 100000 & \multicolumn{1}{r}{20M} & 0.1 \\ 
         & & COD-RNA & 60000 & 8 & 0.1 \\ \hline
        \multirow{2}{*}{Song et al.~\cite{song2013stochastic}} & \multirow{2}{*}{Gradient} & KDDCup99 & 50000 & 9 & 1.0 \\ 
         & & MNIST$^{\dagger}$ & 60000 & 15 & 1.0 \\ \hline
        \multirow{2}{*}{Wu et al.~\cite{wu2017bolt}} & \multirow{2}{*}{Output} & Protein & 72876 & 74 & 0.05 \\ 
         & & CoverType & 498010 & 54 & 0.05 \\ \hline
        \multirow{2}{*}{Jayaraman et al.~\cite{jayaraman2018distributed}} & \multirow{2}{*}{Output} & Adult & 45220 & 104 & 0.5 \\ 
         & & KDDCup99 & 70000 & 122 & 0.5 \\ \hline
    \end{tabular}
    \caption{Simple ERM Methods which achieve High Utility with Low Privacy Budget.} 
    \subcap{$\dagger$ While MNIST is normally a 10-class task, Song et al.~\cite{song2013stochastic} use this for `1 vs rest' binary classification.}
    \label{tab:binaryclassification}
\end{table*}

This section surveys how differential privacy has been used in machine learning applications, with a particular focus on the compromises implementers have made to obtain satisfactory utility. While the effective privacy provided by differential privacy mechanisms depends crucially on the choice of privacy budget $\epsilon$, setting the $\epsilon$ value is discretionary and often done as necessary o achieve acceptable utlity without any consideration of privacy.

Some of the early data analytics works on frequent pattern mining~\cite{bhaskar2010discovering, li2012privbasis}, decision trees~\cite{friedman2010data}, private record linkage~\cite{inan2010private} and recommender systems~\cite{mcsherry2009differentially} were able to achieve both high utility and privacy with $\epsilon$ settings close to 1. These methods rely on finding frequency counts as a sub-routine, and hence provide $\epsilon$-differential privacy by either perturbing the counts using Laplace noise or by releasing the top frequency counts using the exponential mechanism~\cite{mcsherry2007mechanism}. Machine learning, on the other hand, performs much more complex data analysis, and hence requires higher privacy budgets to maintain utility. 

Next, we cover simple binary classification works that use small privacy budgets ($\epsilon \le 1$). Then we survey complex classification tasks which seem to require large privacy budgets. Finally, we summarize recent works that aim to perform complex tasks with low privacy budgets by using variants of differential privacy that offer tighter bounds on composition.

\shortsection{Binary classification}
The first practical implementation of a private machine learning algorithm was proposed by Chaudhuri and Monteleoni~\cite{chaudhuri2009privacy}. They provide a novel sensitivity analysis under strong convexity constraints, allowing them to use output and objective perturbation for binary logistic regression. Chaudhuri et al.~\cite{chaudhuri2011differentially} subsequently generalized this method for ERM algorithms. This sensitivity analysis method has since been used by many works for binary classification tasks under different learning settings (listed in Table~\ref{tab:binaryclassification}). While these applications can be implemented with low privacy budgets ($\epsilon \le 1$), they only perform learning in restricted settings such as learning with low dimensional data, smooth objective functions and strong convexity assumptions, and are only applicable to simple binary classification tasks. 

\begin{table*}[ptb]
    \centering
    \ra{1.1}
    \begin{tabular}{ccccS[table-format=6.0]S[table-format=4.0]S[table-format=2.0]S[table-format=6.0]}
        \toprule
         & Task & Perturbation & Data Set & \multicolumn{1}{c}{$n$} & \multicolumn{1}{c}{$d$} & \multicolumn{1}{c}{$C$} & \multicolumn{1}{c}{$\epsilon$} \\ \hline
       \multirow{2}{*}{Jain et al.~\cite{jain2011differentially}} & \multirow{2}{*}{Online ERM} & \multirow{2}{*}{Objective} & Year & 500000 & 90 & 2 & 10 \\ 
         & & & CoverType & 581012 & 54 & 2 & 10 \\ \hline
        \multirow{5}{*}{Iyengar et al.~\cite{iyengartowards}} & Binary ERM & \multirow{5}{*}{Objective} & Adult & 45220 & 104 & 2 & 10 \\ 
          & Binary ERM & & KDDCup99 & 70000 & 114 & 2 & 10 \\
         & Multi-Class ERM & & CoverType & 581012 & 54 & 7 & 10 \\
         & Multi-Class ERM & & MNIST & 65000 & 784 & 10 & 10 \\
         & High Dimensional ERM & & Gisette & 6000 & 5000 & 2 & 10 \\ \hline
        \multirow{2}{*}{Phan et al.~\cite{phan2016differential, phan2017preserving}} & \multirow{2}{*}{Deep Learning} & \multirow{2}{*}{Objective} & YesiWell & 254 & 30 & 2 & 1 \\ 
         & & & MNIST & 60000 & 784 & 10 & 1 \\ \hline
        \multirow{2}{*}{Shokri and Shmatikov~\cite{shokri2015privacy}} & \multirow{2}{*}{Deep Learning} & \multirow{2}{*}{Gradient} & MNIST & 60000 & 1024 & 10 & 369200 \\
         & & & SVHN & 100000 & 3072 & 10 & 369200 \\ \hline
        \multirow{2}{*}{Zhao et al.~\cite{zhao2018privacy}} & \multirow{2}{*}{Deep Learning} & \multirow{2}{*}{Gradient} & US & 500000 & 20 & 2 & 100 \\ 
         & & & MNIST & 60000 & 784 & 10 & 100 \\ \hline
    \end{tabular}
    \caption{Classification Methods for Complex Tasks}
    \label{tab:higherbudget}
\end{table*}

There has also been considerable progress in generalizing privacy-preserving machine learning to more complex scenarios such as learning in high-dimensional settings~\cite{jain2013differentially, jain2014near, talwar2014private}, learning without strong convexity assumptions~\cite{talwar2015nearly}, or relaxing the assumptions on data and objective functions~\cite{smith2013differentially, zhang2017efficient, wang2017PrivateERMRevisited}. However, these advances are mainly of theoretical interest and only a few works provide implementations~\cite{jain2013differentially, jain2014near}. 

\begin{table*}[ptb]
    \centering
\ra{1.1}
    \begin{tabular}{ccccS[table-format=7.0]S[table-format=4.0]S[table-format=2.0]S[table-format=2.2]}
        \toprule
         & Task & DP Notion & Data Set & \multicolumn{1}{c}{$n$} & \multicolumn{1}{c}{$d$} & \multicolumn{1}{c}{$C$} & \multicolumn{1}{c}{$\epsilon$} \\ \hline
        Huang et al.~\cite{huang2018dp} & ERM & MA & Adult & 21000 & 14 & 2 & 0.5 \\ \hline
        \multirow{2}{*}{Jayaraman et al.~\cite{jayaraman2018distributed}} & \multirow{2}{*}{ERM} & \multirow{2}{*}{zCDP} & Adult & 45220 & 104 & 2 & 0.5\\ 
         & & & KDDCup99 & 70000 & 122 & 2 & 0.5 \\ \hline
        \multirow{4}{*}{Park et al.~\cite{park2017dp}} & \multirow{4}{*}{ERM} & \multirow{4}{*}{zCDP and MA} & Stroke & 50345 & 100 & 2 & 0.5 \\ 
         & & & LifeScience & 26733 & 10 & 2 & 2.0 \\ 
         & & & Gowalla & 1256384 & 2 & 2 & 0.01 \\ 
         & & & OlivettiFace & 400 & 4096 & 2 & 0.3 \\ \hline
        \multirow{3}{*}{Lee~\cite{lee2017differentially}} & \multirow{3}{*}{ERM} & \multirow{3}{*}{zCDP} & Adult & 48842 & 124 & 2 & 1.6 \\ 
         & & & US & 40000 & 58 & 2 & 1.6 \\ 
         & & & Brazil & 38000 & 53 & 2 & 1.6 \\ \hline
        \multirow{3}{*}{Geumlek et al.~\cite{geumlek2017renyi}} & \multirow{3}{*}{ERM} & \multirow{3}{*}{RDP} & Abalone & 2784 & 9 & 2 & 1.0 \\ 
         & & & Adult & 32561 & 100 & 2 & 0.05 \\ 
         & & & MNIST & 7988 & 784 & 2 & 0.14 \\ \hline
        \multirow{2}{*}{Beaulieu et al.~\cite{beaulieu2018privacy}} & \multirow{2}{*}{Deep Learning} & \multirow{2}{*}{MA} & eICU & 4328 & 11 & 2 & 3.84 \\ 
         & & & TCGA & 994 & 500 & 2 & 6.11 \\ \hline
        \multirow{2}{*}{Abadi et al.~\cite{abadi2016deep}} & \multirow{2}{*}{Deep Learning} & \multirow{2}{*}{MA} & MNIST & 60000 & 784 & 10 & 2.0 \\ 
         & & & CIFAR & 60000 & 3 072 & 10 & 8.0 \\ \hline
        \multirow{2}{*}{Yu et al.~\cite{yudifferentially}} & \multirow{2}{*}{Deep Learning} & \multirow{2}{*}{MA} & MNIST & 60000 & 784 & 10 & 21.5 \\ 
         & & & CIFAR & 60000 & 3072 & 10 & 21.5 \\ \hline 
        \multirow{2}{*}{Papernot et al.~\cite{papernot2017semi}} & \multirow{2}{*}{Deep Learning} & \multirow{2}{*}{MA} & MNIST & 60000 & 784 & 10 & 2.0 \\ 
         & & & SVHN & 60000 & 3072 & 10 & 8.0 \\ \hline
        Geyer et al.~\cite{geyer2017differentially} & Deep Learning & MA & MNIST & 60000 & 784 & 10 & 8.0 \\ \hline
        \multirow{2}{*}{Bhowmick et al.~\cite{bhowmick2018protection}} & \multirow{2}{*}{Deep Learning} & \multirow{2}{*}{MA} & MNIST & 60000 & 784 & 10 & 3.0 \\ 
         & & & CIFAR & 60000 & 3072 & 10 & 3.0 \\ \hline
        Hynes et al.~\cite{hynes2018efficient} & Deep Learning & MA & CIFAR & 50000 & 3072 & 10 & 4.0 \\ \hline
    \end{tabular}
    \caption{Gradient Perturbation based Classification Methods using Different Notions of Differential Privacy}
    \label{tab:relaxednotions}
\end{table*}

\shortsection{Complex learning tasks}
All of the above works are limited to convex learning problems with binary classification tasks. Adopting their approaches to more complex learning tasks requires higher privacy budgets (see Table~\ref{tab:higherbudget}). For instance, the online version of ERM as considered by Jain et al.~\cite{jain2011differentially} requires $\epsilon$ as high as 10 to achieve acceptable utility. From the definition of differential privacy, we can see that  $\text{Pr}[\mathcal{M}(D) \in S] \le e^{10} \times \text{Pr}[\mathcal{M}(D') \in S]$. In other words, even if the model's output probability is 0.0001 on a data set $D'$ that doesn't contain the target record, the model's output probability can be as high as 0.9999 on a neighboring data set $D$ that contains the record. This allows an adversary to infer the presence or absence of a target record from the training data with high confidence. Adopting these binary classification methods for multi-class classification tasks requires even higher $\epsilon$ values. As noted by Wu et al.~\cite{wu2017bolt}, it would require training a separate binary classifier for each class. 
Finally, high privacy budgets are required for non-convex learning algorithms, such as deep learning~\cite{shokri2015privacy, zhao2018privacy}. Since the output and objective perturbation methods of Chaudhuri et al.~\cite{chaudhuri2011differentially} are not applicable to non-convex settings, implementations of differentially private deep learning rely on gradient perturbation in their iterative learning procedure. These methods do not scale to large numbers of training iterations due to the composition theorem of differential privacy which causes the privacy budget to accumulate across iterations. The only exceptions are the works of Phan et al.~\cite{phan2016differential, phan2017preserving} that replace the non-linear functions in deep learning with polynomial approximations and then apply objective perturbation. With this transformation, they achieve high model utility for $\epsilon = 1$, as shown in Table~\ref{tab:higherbudget}. However, we note that this polynomial approximation is a non-standard approach to deep learning which can limit the model's learning capacity, and thereby diminish the model's accuracy for complex tasks. 

\shortsection{Machine learning with other DP definitions}
To avoid the stringent composition property of differential privacy, several proposed privacy-preserving deep learning methods adopt the differential privacy variants introduced in  Section~\ref{relaxed_definitions}. 
Table~\ref{tab:relaxednotions} lists works that use gradient perturbation with differential privacy variants to reduce the overall privacy budget during iterative learning. The utility benefit of using these variants is evident from the fact that the privacy budget for deep learning algorithms is significantly less than the prior works of Shokri and Shmatikov~\cite{shokri2015privacy} and Zhao et al.~\cite{zhao2018privacy}. 

While these variants of differential privacy make complex iterative learning feasible for reasonable $\epsilon$ values, they might lead to more privacy leakage in practice. The main goal of our study is to evaluate the impact of implementation decisions regarding the privacy budget and variants of differential privacy on the concrete privacy leakage that can be exploited by an attacker in practice. We do this by experimenting with various inference attacks, described in the next section.

\section{Inference Attacks on Machine Learning}\label{sec:attacks}

This section surveys the two types of inference attacks, \emph{membership inference} (Section~\ref{sec:membershipinference}) and \emph{attribute inference} (Section~\ref{sec:attributeinference}), and explains why they are useful metrics for evaluating privacy leakage. Section~\ref{sec:otherattacks} briefly summarizes other relevant privacy attacks on machine learning.

\subsection{Membership Inference}\label{sec:membershipinference}
The aim of a \emph{membership inference} attack is to infer whether or not a given record is present in the training set. Membership inference attacks can uncover highly sensitive information from training data. An early membership inference attack showed that it is possible to identify individuals contributing DNA to studies that analyze a mixture of DNA from many individuals, using a statistical distance measure to determine if a known individual is in the mixture~\cite{homer2008resolving}. 

Membership inference attacks can either be completely black-box where an attacker only has query access to the target model~\cite{shokri2017membership}, or can assume that the attacker has full white-box access to the target model, along with some auxiliary information~\cite{yeom2018privacy}. 
The first membership inference attack on machine learning was proposed by Shokri et al.~\cite{shokri2017membership}.  They consider an attacker who can query the target model in a black-box way to obtain confidence scores for the queried input. The attacker tries to exploit the confidence score to determine whether the query input was present in the training data. Their attack method involves first training shadow models on a labelled data set, which can be generated either via black-box queries to the target model or through assumptions about the underlying distribution of training set. The attacker then trains an attack model using the shadow models to distinguish whether or not an input record is in the shadow training set. Finally, the attacker makes API calls to the target model to obtain confidence scores for each given input record and infers whether or not the input was part of the target model's training set. The inference model distinguishes the target model’s predictions for inputs that are in its training set from those it did not train on. The key assumption is that the confidence score of the target model is higher for the training instances than it would be for arbitrary instances not present in the training set. This can be due to the generalization gap, which is prominent in models that overfit to training data.

A more targeted approach was proposed by Long et al.~\cite{long2017towards} where the shadow models are trained with and without a targeted input record $t$. At inference time, the attacker can check if the input record $t$ was present in the training set of target model. This approach tests the membership of a specific record more accurately than Shokri et al.'s approach~\cite{shokri2017membership}. Recently, Salem et al.~\cite{salem2019ml} proposed more generic membership inference attacks by relaxing the requirements of Shokri et al.~\cite{shokri2017membership}. In particular, requirements on the number of shadow models, knowledge of training data distribution and the target model architecture can be relaxed without substantially degrading the effectiveness of the attack.

Yeom et al.~\cite{yeom2018privacy} recently proposed a more computationally efficient membership inference attack when the attacker has access to the target model and knows the average training loss of the model. To test the membership of an input record, the attacker evaluates the loss of the model on the input record and then classifies it as a member if the loss is smaller than the average training loss. 

\shortsection{Connection to Differential Privacy} Differential privacy, by definition, aims to obfuscate the presence or absence of a record in the data set. On the other hand, membership inference attacks aim to identify the presence or absence of a record in the data set. Thus, intuitively these two notions counteract each other. Li et al.~\cite{li2013membership} point to this fact and provide a direct relationship between differential privacy and membership inference attacks. Backes et al.~\cite{backes2016membership} studied membership inference attacks on microRNA studies and showed that differential privacy can reduce the success of membership inference attacks, but at the cost of utility. 

Yeom et al.~\cite{yeom2018privacy} formally define a membership inference attack as an adversarial game where a data element is selected from the distribution, which is randomly either included in the training set or not. Then, an adversary with access to the trained model attempts to determine if that element was used in training. The \emph{membership advantage} is defined as the difference between the adversary's true and false positive rates for this game. The authors prove that if the learning algorithm satisfies $\epsilon$-differential privacy, then the adversary's advantage is bounded by $e^\epsilon - 1$. 
Hence, it is natural to use membership inference attacks as a metric to evaluate the privacy leakage of differentially private algorithms.

\subsection{Attribute Inference}\label{sec:attributeinference}
The aim of an \emph{attribute inference} attack (also called \emph{model inversion}) is to learn hidden sensitive attributes of a test input given at least API access to the model and information about the non-sensitive attributes. Fredrikson et al.~\cite{fredrikson2014privacy} formalize this attack in terms of maximizing the posterior probability estimate of the sensitive attribute. More concretely, for a test record $x$ where the attacker knows the values of its non-sensitive attributes $x_1, x_2, \cdots x_{d-1}$ and all the prior probabilities of the attributes, the attacker obtains the output of the model, $f(x)$, and attempts to recover the value of the sensitive attribute $x_d$. The attacker essentially searches for the value of $x_d$ that maximizes the posterior probability $P(x_d\; | \; x_1, x_2, \cdots x_{d-1}, f(x))$.  The success of this attack is based on the correlation between the sensitive attribute, $x_d$, and the model output, $f(x)$. 

Yeom et al.~\cite{yeom2018privacy} also propose an attribute inference attack using the same principle they use for their membership inference attack. The attacker evaluates the model's empirical loss on the input instance for different values of the sensitive attribute, and reports the value which has the maximum posterior probability of achieving the empirical loss. The authors define the \emph{attribute advantage} similarly to their definition of membership advantage for membership inference.

Fredrikson et al.~\cite{fredrikson2014privacy} demonstrated attribute inference attacks that could identify genetic markers based on warfarin dosage output by a model with just black-box access to model API.\footnote{This application has stirred some controversy based on the warfarin dosage output by the model itself being sensitive information correlated to the sensitive genetic markers, hence the assumption on attacker's prior knowledge of warfarin dosage is somewhat unrealistic~\cite{mcsherry2016warfarin}.} With additional access to confidence scores of the model (noted as white-box information by Wu et al.~\cite{wu2016methodology}), more complex tasks have been performed, such as recovering faces from the training data~\cite{fredrikson2015model}. 

\shortsection{Connection to Differential Privacy}
Differential privacy is mainly tailored to obfuscate the presence or absence of a record in a data set, by limiting the effect of any single record on the output of differential private model trained on the data set. Logically this definition also extends to attributes or features of a record. In other words, by adding sufficient differential privacy noise, we should be able to limit the effect of a sensitive attribute on the model's output. This relationship between records and attributes is discussed by Yeom et al.~\cite{yeom2018privacy}. Hence, we include these attacks in our experiments.

\subsection{Other Attacks on Machine Learning}\label{sec:otherattacks}
Apart from inference attacks, many other attacks have been proposed in the literature which try to infer specific information from the target model. 
The most relevant are memorization attacks, which try to exploit the ability of high capacity models to memorize certain sensitive patterns in the training data~\cite{carlini2018secret}.  These attacks have been found to be thwarted by differential privacy mechanisms with very little noise ($\epsilon = 10^9$)~\cite{carlini2018secret}. 

Other privacy attacks include model stealing, hyperparameter stealing, and property inference attacks. A model stealing attack aims to recover the model parameters via black-box access to the target model, either by adversarial learning~\cite{lowd2005adversarial} or by equation solving attacks~\cite{tramer2016stealing}.
Hyperparameter stealing attacks try to recover the underlying hyperparameters used during the model training, such as regularization coefficient~\cite{wang2018stealing} or model architecture~\cite{yan2018cache}. These hyperparameters are intellectual property of commercial organizations that deploy machine learning models as a service, and hence these attacks are regarded as a threat to valuable intellectual property. A property inference attack tries to infer whether the training data set has a specific property, given a white-box access to the trained model. For instance, given access to a speech recognition model, an attacker can infer if the training data set contains speakers with a certain accent. Here the attacker can use the shadow training method of Shokri et al.~\cite{shokri2017membership} for distinguishing the presence and absence of a target property. These attacks have been performed on HMM and SVM models~\cite{ateniese2015hacking} and neural networks~\cite{ganju2018property}. 

Though all these attacks may leak sensitive information about the target model or training data, the information leaked tends to be application-specific and is not clearly defined in a general way. For example, a property inference attack leaks some statistical property of the training data that is surprising to the model developer. Of course, the overall purpose of the model is to learn statistical properties from the training data. So, there is no general definition of a property inference attack without a prescriptive decision about which statistical properties of the training data should be captured by the model and which are sensitive to leak. In addition, the attacks mentioned in this section do not closely follow the threat model of differential privacy. Thus, we only consider inference attacks for our experimental evaluation. 

In addition to these attacks, several attacks have been proposed that require an adversary that can actively interfere with the model training process~\cite{xiao2015support, Munoz-Gonzalez:2017:TPD:3128572.3140451, DBLP:journals/corr/abs-1807-00459, DBLP:journals/corr/YangWLC17}. We consider these out of scope, and assume a clean training process not under the control of the adversary.

\section{Empirical Evaluation}\label{sec:experiments}

To quantify observable privacy leakage, we conduct experiments to measure how much an adversary can infer from a model. As motivated in Section~\ref{sec:attacks}, we measure privacy leakage using membership and attribute inference in our experiments. 
Note, however, that the conclusions we can draw from experiments like this are limited to showing a \emph{lower bound} on the information leakage since they are measuring the effectiveness of a particular attack.
This contrasts with differential privacy guarantees, which provide an \emph{upper bound} on possible leakage. Experimental results cannot be used to make strong claims about what the best possible attack would be able to infer, especially in cases where an adversary has auxiliary information to help guide the attack. Evidence from our experiments, however, can provide clear evidence that implemented privacy protections do not appear to provide sufficient privacy.

\subsection{Experimental Setup}\label{sec:ex:setup}

We evaluate the privacy leakage of two differentially private algorithms using gradient perturbation: logistic regression for empirical risk minimization (Section~\ref{sec:results:erm}) and neural networks for non-convex learning (Section~\ref{sec:results:nn}). 
For both, we consider the different notions of differential privacy and compare their privacy leakage. The variations that we implement are na\"{i}ve composition (NC), advanced composition (AC), zero-concentrated differential privacy (zCDP) and R\'{e}nyi differential privacy (RDP) (see Section~\ref{relaxed_definitions} for details). We do not include CDP as it has the same composition property as zCDP (Table~\ref{tab:dp_variations_comparison}). 
For RDP, we use the RDP accountant (previously known as the moments accountant) of the TF Privacy framework~\cite{tf-privacy}.

We evaluate the models on two main metrics: \emph{accuracy loss}, the model's accuracy on a test set relative to the non-private baseline model, and \emph{privacy leakage}, the attacker's advantage as defined by Yeom et al.~\cite{yeom2018privacy}. The accuracy loss is normalized with respect to the accuracy of non-private model to clearly depict the model utility: 
\[\text{\em Accuracy Loss} = 1 - \frac{\text{\em Accuracy of Private Model}}{\text{\em Accuracy of Non-Private Model}}.\]

To evaluate the inference attack, we provide the attacker with a set of 20,000 records consisting of 10,000 records from training set and 10,000 records from the test set. We call records in the training set \emph{members}, and the other records \emph{non-members}. These labels are not known to the attacker. The task of the attacker is to predict whether or not a given input record belongs to the training set (i.e., if it is a member). The privacy leakage metric is calculated by taking the difference between the true positive rate (TPR) and the false positive rate (FPR) of the inference attack. Thus the privacy leakage metric is always between 0 and 1, where the value of 0 indicates that there is no leakage. For example, given that there are 100 member and 100 non-member records, if an attacker performs membership inference on the model and correctly identifies all the `true' member records while falsely labelling 30 non-member records as members, then the privacy leakage would be 0.7. To better understand the potential impact of leakage, we also conduct experiments to estimate the actual number of members who are at risk for disclosure in a membership inference attack.

\shortsection{Data sets}
We evaluate our models over two data sets for multi-class classification tasks: \dataset{CIFAR-100}~\cite{krizhevsky2009learning} and \dataset{Purchase-100}~\cite{purchase}. 
\dataset{CIFAR-100} consists of $28 \times 28$ images of 100 real world objects, with 500 instances of each object class. We use PCA to reduce the dimensionality of records to 50. 
The \dataset{Purchase-100} data set consists of 200,000 customer purchase records of size 100 each (corresponding to the 100 frequently-purchased items) where the records are grouped into 100 classes based on the customers' purchase style. For both data sets, we use 10,000 randomly-selected instances for training and 10,000 randomly-selected non-training instances for the test set. The remaining records are used for training the shadow models and inference model used in the Shokri et al.\ attacks.

\shortsection{Attacks}
For our experiments, we use the attack frameworks of Shokri et al.~\cite{shokri2017membership} and Yeom et al.~\cite{yeom2018privacy} for membership inference and the method proposed by Yeom et al.~\cite{yeom2018privacy} for attribute inference. 
In Shokri et al.'s framework~\cite{shokri2017membership}, multiple shadow models are trained on data that is sampled from the same distribution as the private data set. These shadow models are used to train an inference model to identify whether an input record belongs to the private data set. The inference model is trained using a set of records used to train the shadow models, a set of records randomly selected from the distribution that are not part of the shadow model training, along with the confidence scores output by the shadow models for all of the input records. Using these inputs, the inference model learns to distinguish the training records from the non-training records. At the inference stage, the inference model takes an input record along with the confidence score of the target model on the input record, and outputs whether the input record belongs to the target model's private training data set. The intuition is that if the target model overfits on its training set, its confidence score for a training record will be higher than its confidence score for an otherwise similar input that was not used in training. The inference model tries to exploit this property. In our instantiation of the attack framework, we use five shadow models which all have the same model architecture as the target model. Our inference model is a neural network with two hidden layers of size 64. This setting is consistent with the original work~\cite{shokri2017membership}. 

The attack framework of Yeom et al.~\cite{yeom2018privacy} is simpler than Shokri et al.'s design. It assumes the attacker has access to the target model's expected training loss on the private training data set, in addition to having access to the target model. For membership inference, the attacker simply observes the target model's loss on the input record. The attacker classifies the record as a member if the loss is smaller than the target model's expected training loss, otherwise the record is classified as a non-member. The same principle is used for attribute inference. Given an input record, the attacker brute-forces all possible values for the unknown attribute and observes the target model's loss, outputting the value for which the loss is closest to the target's expected training loss. Since there are no attributes in our data sets that are explicitly annotated as private, we randomly choose five attributes, and perform the attribute inference attack on each attribute independently, and report the averaged results.

\shortsection{Hyperparameters}
For both data sets, we train logistic regression and neural network models with $\ell_2$ regularization. First, we train a non-private model and perform a grid search over the regularization coefficient $\lambda$ to find the value that minimizes the classification error on the test set. For \dataset{CIFAR-100}, we found optimal values to be $\lambda = 10^{-5}$ for logistic regression and $\lambda = 10^{-4}$ for neural network. For \dataset{Purchase-100}, we found optimal values to be $\lambda = 10^{-5}$ for logistic regression and $\lambda = 10^{-8}$ for neural network. Next, we fix this setting to train differentially private models using gradient perturbation. We vary $\epsilon$ between 0.01 and 1000 while keeping $\delta = 10^{-5}$, and report the accuracy loss and privacy leakage. The choice of $\delta = 10^{-5}$ satisfies the requirement that $\delta$ should be smaller than the inverse of the training set size 10,000. We use the ADAM optimizer for training and fix the learning rate to 0.01 with a batch size of 200. Due to the random noise addition, all the experiments are repeated five times and the average results and standard errors are reported. We do not assume pre-trained model parameters, unlike the prior works of Abadi et al.~\cite{abadi2016deep} and Yu et al.~\cite{yudifferentially}.

\begin{figure*}[tb]
\centering
    \subfigure[Batch gradient clipping]{
    \includegraphics[width=0.45\textwidth]{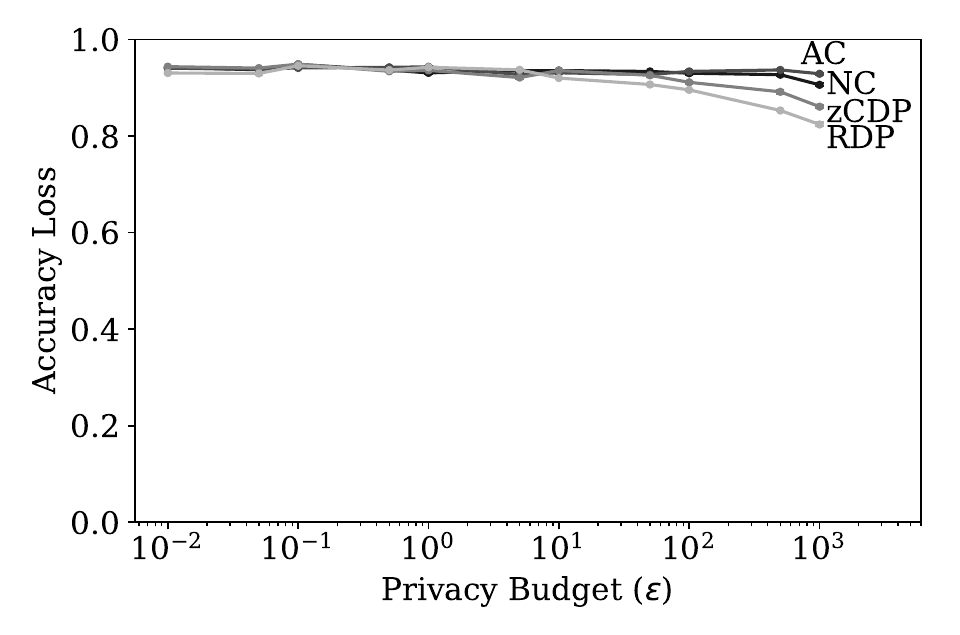}
    \label{fig:cifar_lr_grad_acc_batch}}
    \subfigure[Per-instance gradient clipping]{
    \includegraphics[width=0.45\textwidth]{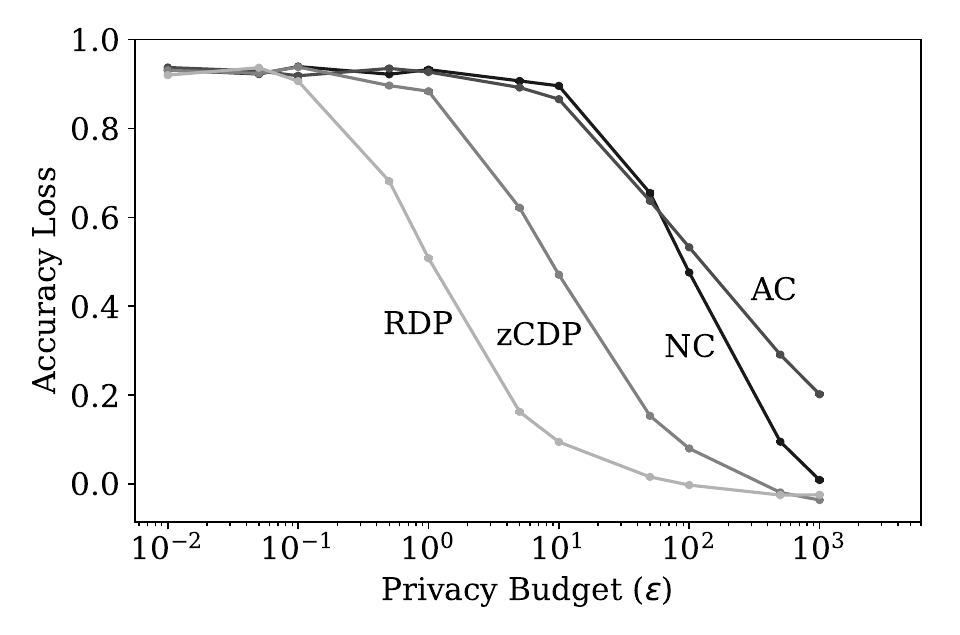}
    \label{fig:cifar_lr_grad_acc_instance}}
\caption{Impact of clipping on accuracy loss of logistic regression (\dataset{CIFAR-100}).}
\label{fig:cifar_lr_grad_acc}
\end{figure*}

\shortsection{Clipping} 
For gradient perturbation, clipping is required to bound the sensitivity of the gradients. We tried clipping at both the batch and per-instance level. Batch clipping is more computationally efficient and a standard practice in deep learning. On the other hand, per-instance clipping uses the privacy budget more efficiently, resulting in more accurate models for a given privacy budget. We use the TensorFlow Privacy framework~\cite{tf-privacy} which implements both batch and per-instance clipping. We fix the clipping threshold at $\cC = 1$. 

Figure \ref{fig:cifar_lr_grad_acc} compares the accuracy loss of logistic regression models trained over \dataset{CIFAR-100} data set with both batch clipping and per-instance clipping. Per-instance clipping allows learning more accurate models for all values of $\epsilon$ and amplifies the differences between the different mechanisms. For example, while none of the models learn anything useful when using batch clipping, the model trained with RDP achieves accuracy close to the non-private model for $\epsilon = 100$ when performing per-instance clipping. Hence, for the rest of the paper we only report the results for per-instance clipping.

\begin{figure*}[ptb]
\centering
    \subfigure[Shokri et al.\ membership inference]{
    \includegraphics[width=0.32\textwidth]{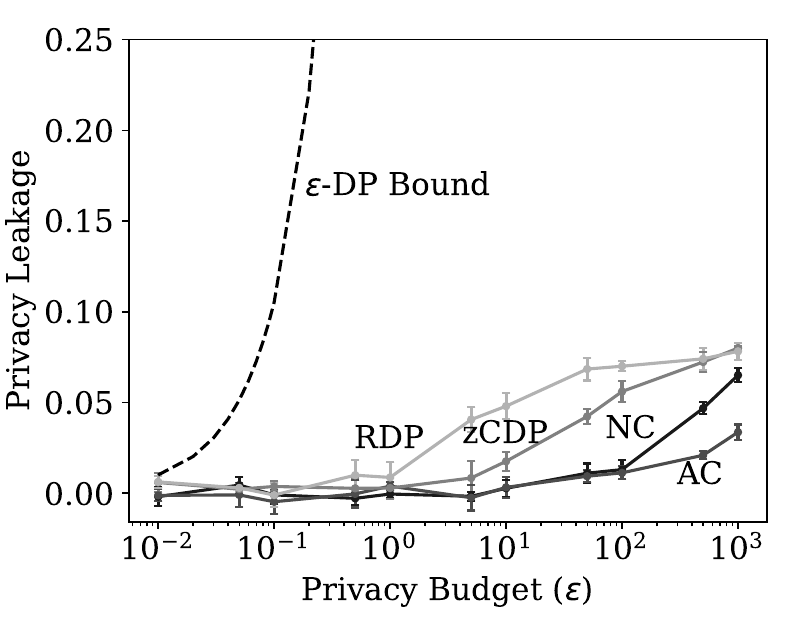}
    \label{fig:cifar_lr_inference_attack}}
    \subfigure[Yeom et al.\ membership inference]{
    \includegraphics[width=0.32\textwidth]{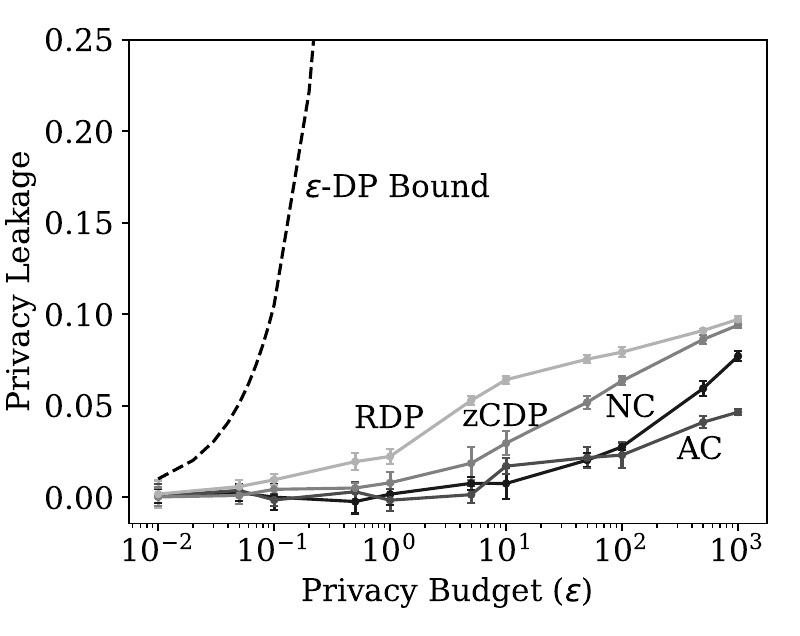}
    \label{fig:cifar_lr_inference_mem}}
    \subfigure[Yeom et al.\ attribute inference]{
    \includegraphics[width=0.32\textwidth]{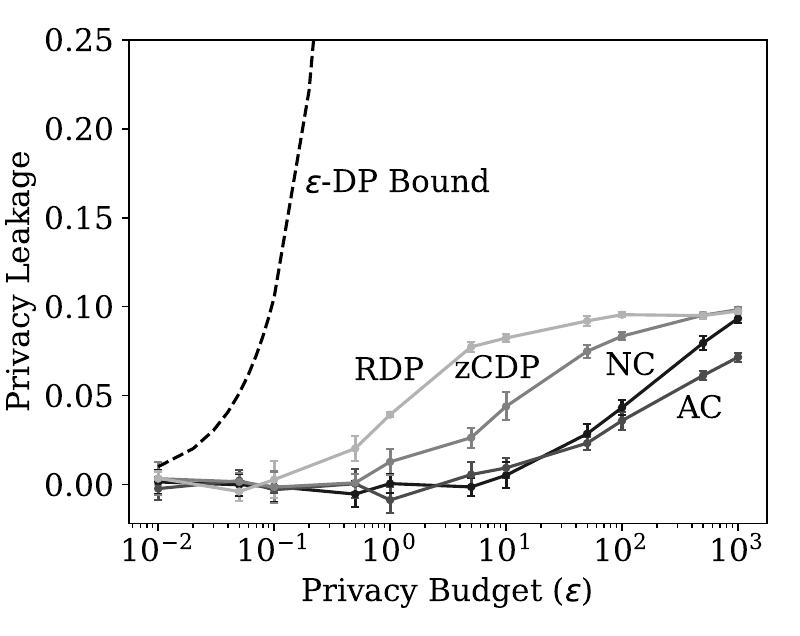}
    \label{fig:cifar_lr_inference_attr}}
\caption{Inference attacks on logistic regression (\dataset{CIFAR-100}).}
\label{fig:cifar_lr_inference}
\end{figure*}

\subsection{Logistic Regression Results}\label{sec:results:erm}

We train $\ell_2$-regularized logistic regression models on both the \dataset{CIFAR-100} and \dataset{Purchase-100} data sets.

\shortsection{\bfdataset{CIFAR-100}}
The baseline model for non-private logistic regression achieves accuracy of 0.225 on training set and 0.155 on test set, which is competitive with the state-of-art neural network model~\cite{shokri2017membership} that achieves test accuracy close to 0.20 on \dataset{CIFAR-100} after training on larger data set. Thus, there is a small generalization gap of 0.07 for the inference attacks to exploit.

Figure~\ref{fig:cifar_lr_grad_acc_instance} compares the accuracy loss for logistic regression models trained with different notions of differential privacy as we vary the privacy budget $\epsilon$. As depicted in the figure, na\"{i}ve composition achieves accuracy close to 0.01 for $\epsilon \le 10$ which is random guessing for 100-class classification. Na\"{i}ve composition achieves accuracy loss close to 0 for $\epsilon = 1000$. Advanced composition adds more noise than na\"{i}ve composition when the privacy budget is greater than the number of training epochs ($\epsilon \ge 100$), so it should never be used in such settings. The zCDP and RDP variants achieve accuracy loss close to 0 at $\epsilon = 500$ and $\epsilon = 50$ respectively, which is order of magnitudes smaller than the na\"{i}ve composition. This is expected since these variations require less added noise for the same privacy budget. 

Figures~\ref{fig:cifar_lr_inference_attack} and~\ref{fig:cifar_lr_inference_mem} show the privacy leakage due to membership inference attacks on logistic regression models. Figure~\ref{fig:cifar_lr_inference_attack} shows results for the Shokri et al.\ attack~\cite{shokri2017membership}, which has access to the target model's confidence scores on the input record. Na\"{i}ve composition achieves privacy leakage close to 0 for $\epsilon \le 10$, and the leakage reaches $0.065 \pm 0.004$ for $\epsilon = 1000$. The RDP and zCDP variants have average leakage close to $0.080 \pm 0.004$ for $\epsilon = 1000$. As expected, the differential privacy variations have leakage in accordance with the amount of noise they add for a given $\epsilon$.  The plots also include a dashed line showing the theoretical upper bound on the privacy leakage for $\epsilon$-differential privacy, where the bound is $e^\epsilon - 1$ (see Section~\ref{sec:membershipinference}). As depicted, there is a huge gap between the theoretical upper bound and the empirical privacy leakage for all the variants of differential privacy. This implies that more powerful inference attacks could exist in practice.

Figure~\ref{fig:cifar_lr_inference_mem} shows results for the white-box attacker of Yeom et al.~\cite{yeom2018privacy}, which has access to the target model's loss on the input record. 
As expected, zCDP and RDP leak the most. Na\"{i}ve composition does not have any significant leakage for $\epsilon \le 10$, but the leakage reaches $0.077 \pm 0.003$ for $\epsilon = 1000$. The observed leakage of all the variations is in accordance with the noise magnitude required for different differential privacy guarantees. From Figure~\ref{fig:cifar_lr_grad_acc_instance} and Figure~\ref{fig:cifar_lr_inference}, we see that RDP at $\epsilon = 10$ achieves similar utility and privacy leakage to NC at $\epsilon = 500$.

Figure~\ref{fig:cifar_lr_inference_attr} depicts the privacy leakage due to the attribute inference attack. Na\"{i}ve composition has low privacy leakage for $\epsilon \le 10$ (attacker advantage of $0.005 \pm 0.007$ at $\epsilon = 10$), but it quickly increases to $0.093 \pm 0.002$ for $\epsilon = 1000$. 
As expected, across all variations as privacy budgets increase both the attacker's advantage (privacy leakage) and the model utility (accuracy) increase.

\begin{figure}[ptb]
    \centering
    \includegraphics[width=\linewidth]{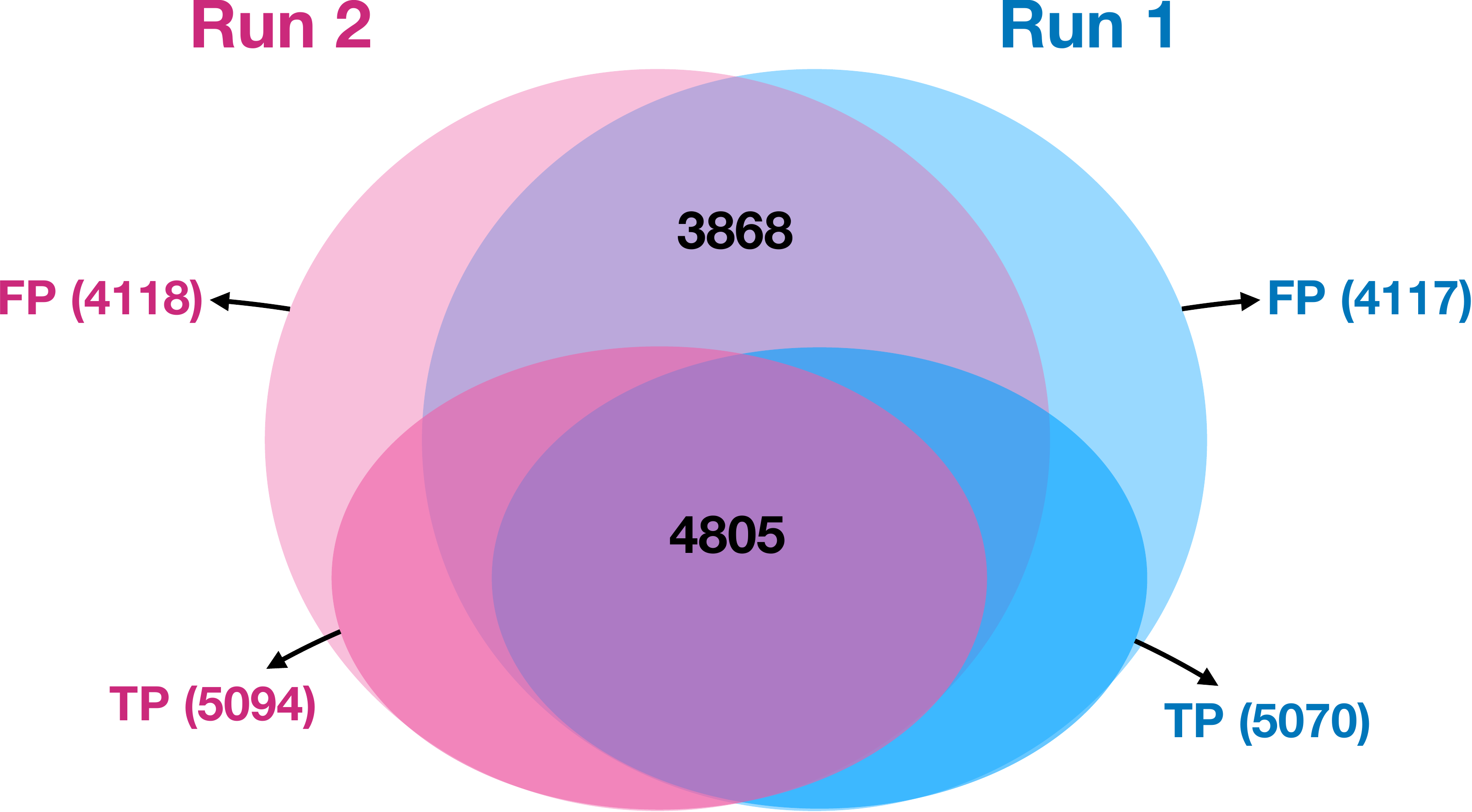}
    \caption{Overlap of membership predictions across two runs of logistic regression with RDP at $\epsilon = 1000$ (\dataset{CIFAR-100})}
    \label{fig:cifar_lr_overlap}
\end{figure}

We further analyze the impact of privacy leakage on individual members, to understand whether exposure is more determined by the random noise added in training or depends on properties of particular records. To do so, we use the membership inference attack of Yeom et al.\ on the models trained with RDP at $\epsilon = 1000$, since this is the setting where the privacy leakage is the highest. As shown in Figure~\ref{fig:cifar_lr_inference_mem}, the average leakage across five runs for this setting is 0.10 where the average positive predictive value\footnote{PPV gives the fraction of correct membership predictions among all the membership predictions made by the attacker. A PPV value of 0.5 means that the adversary has no advantage, i.e. the adversary cannot distinguish between members and non-members. Thus only PPV values greater than 0.5 are of significance to the adversary.} (PPV) is 0.55. On average, the attack identifies 9,196 members, of which 5,084 are true members (out of 10,000 members) and 4,112 are false positives (out of 10,000 non-members). To see if the members are exposed randomly across different runs, we calculate the overlap of membership predictions. Figure~\ref{fig:cifar_lr_overlap} illustrates the overlap of membership predictions across two of the five runs. Out of 9,187 membership predictions, 5,070 members are correctly identified (true positives) and 4,117 non-members are incorrectly classified as members (false positives) in the first run, and 5,094 members are correctly identified in the second run. Across both the runs, there are 8,673 records predicted as members in both runs, of which 4,805 are true members (PPV = 0.55). Thus, the overlap is much higher than would be expected if exposure risks was due only to the random noise added. It depends on properties of the data, although the relatively small increase in PPV suggests they are not highly correlated with actual training set membership.

\begin{figure}[ptb]
    \centering
    \includegraphics[width=\linewidth]{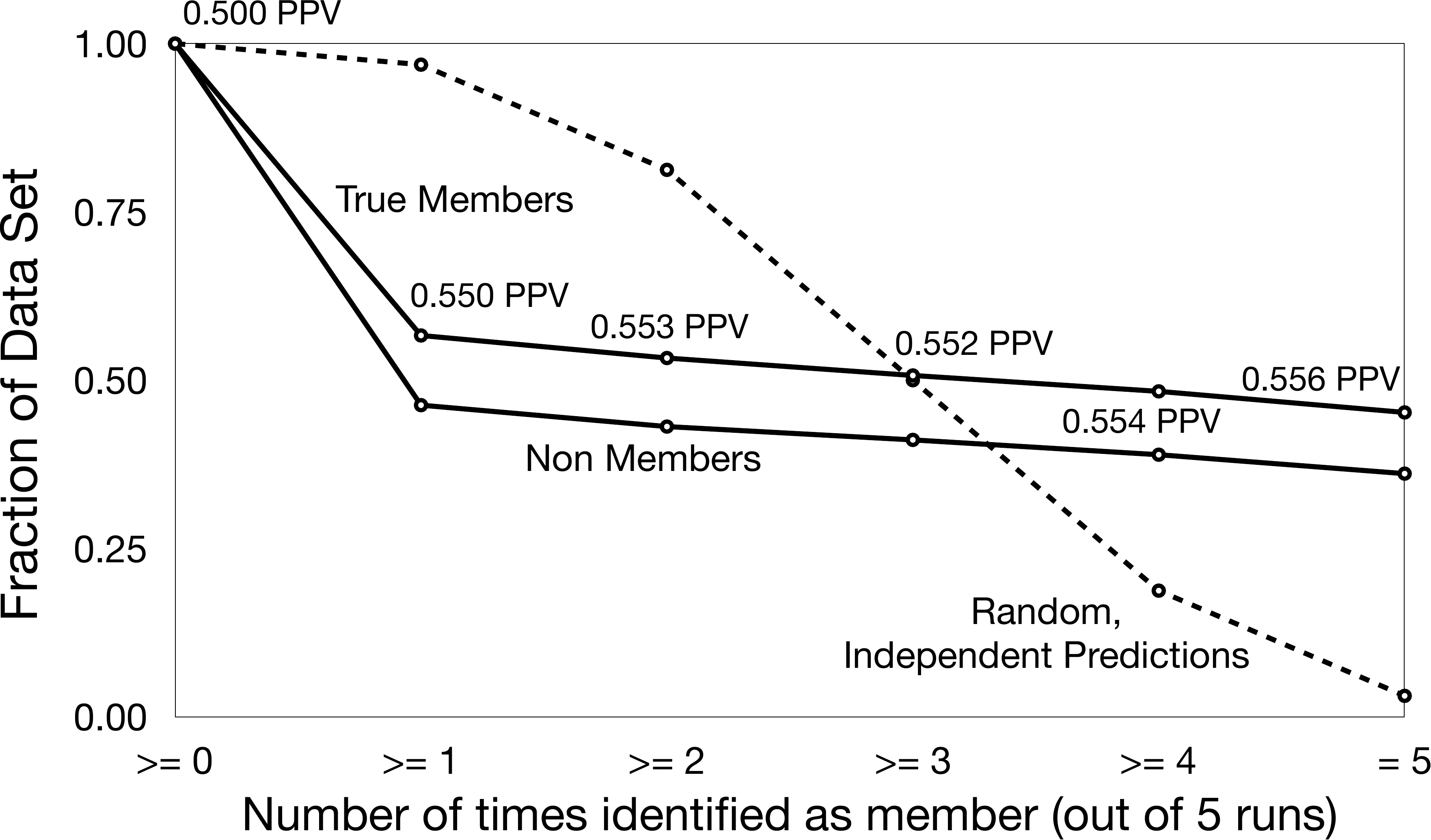}
    \caption{Membership predictions across multiple runs of logistic regression with RDP at $\epsilon = 1000$ (\dataset{CIFAR-100})}
    \label{fig:cifar_lr_multiple_runs}
\end{figure}

\begin{figure}[tb]
    \centering
    \includegraphics[width=.45\textwidth]{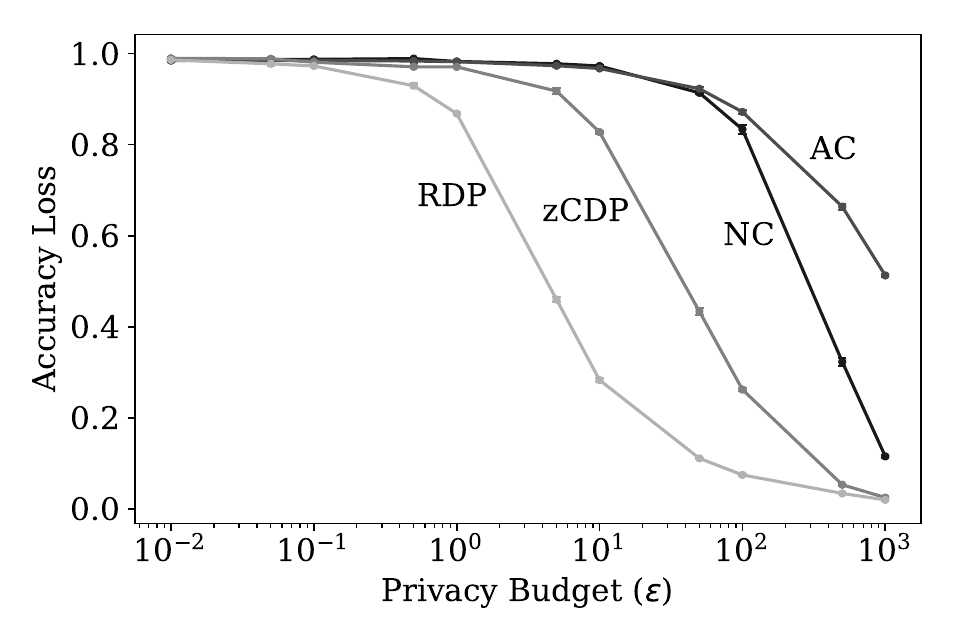}
    \caption{Accuracy loss of logistic regression (\dataset{Purchase-100}).}
    \label{fig:purchase_lr_grad_acc}
\end{figure}

\begin{figure*}[ptb]
\centering
    \subfigure[Shokri et al.\ membership inference]{
    \includegraphics[width=0.32\textwidth]{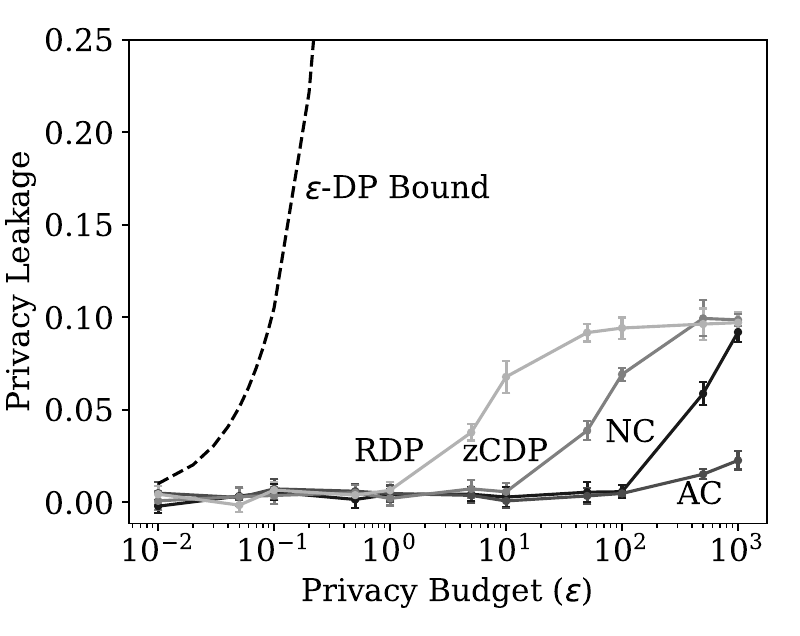}
    \label{fig:purchase_lr_grad_inference_attack}}
    \subfigure[Yeom et al.\ membership inference]{
    \includegraphics[width=0.32\textwidth]{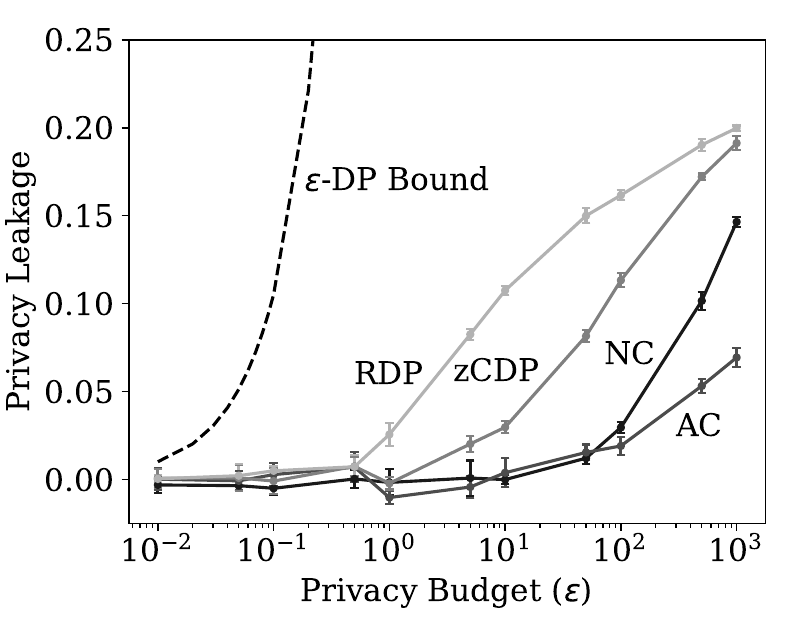}
    \label{fig:purchase_lr_grad_inference_mem}}
    \subfigure[Yeom et al.\ attribute inference]{
    \includegraphics[width=0.32\textwidth]{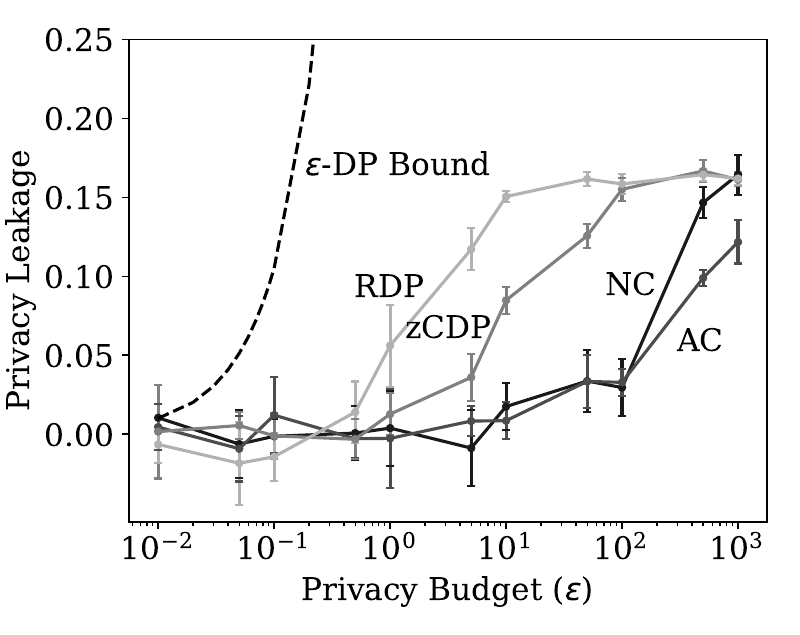}
    \label{fig:purchase_lr_grad_inference_attr}}
\caption{Inference attacks on logistic regression (\dataset{Purchase-100}).}
\label{fig:purchase_lr_grad_inference}
\end{figure*}

Figure~\ref{fig:cifar_lr_multiple_runs} shows how many members and non-members are predicted as members multiple times across five runs.
For comparison, the dotted line depicts the fraction of records that would be identified as members by a test that randomly and independently predicted membership with $0.5$ probability (so, PPV=0.5 at all points).
This result corresponds to a scenario where models are trained independently five times on the same data set, and the adversary has access to all the model releases. This is not a realistic attack scenario, but might approximate a scenario where a model is repeatedly retrained using mostly similar training data. Using this extra information, the adversary can be more confident about predictions that are made consistently across multiple models. As the figure depicts, however, in this case the PPV increases slightly with intersection of more runs but remains close to 0.5.  Thus, even after observing the outputs of all the five models, the adversary has little advantage in this setting. While this seems like the private models are performing well, the adversary identifies 5,179 members with a PPV of 0.56 even with the non-private model. Hence the privacy benefit here is not due to the privacy mechanisms --- even without any privacy noise, the logistic regression model does not appear to leak enough information to enable effective membership inference attacks. This highlights the huge gap between the theoretical upper bound on privacy leakage and the empirical leakage of the implemented inference attacks (Figure~\ref{fig:cifar_lr_inference}), showing that there could be more powerful inference attacks in practice or ways to provide more meaningful privacy guarantees. 

\begin{figure}[ptb]
    \centering
    \includegraphics[width=\linewidth]{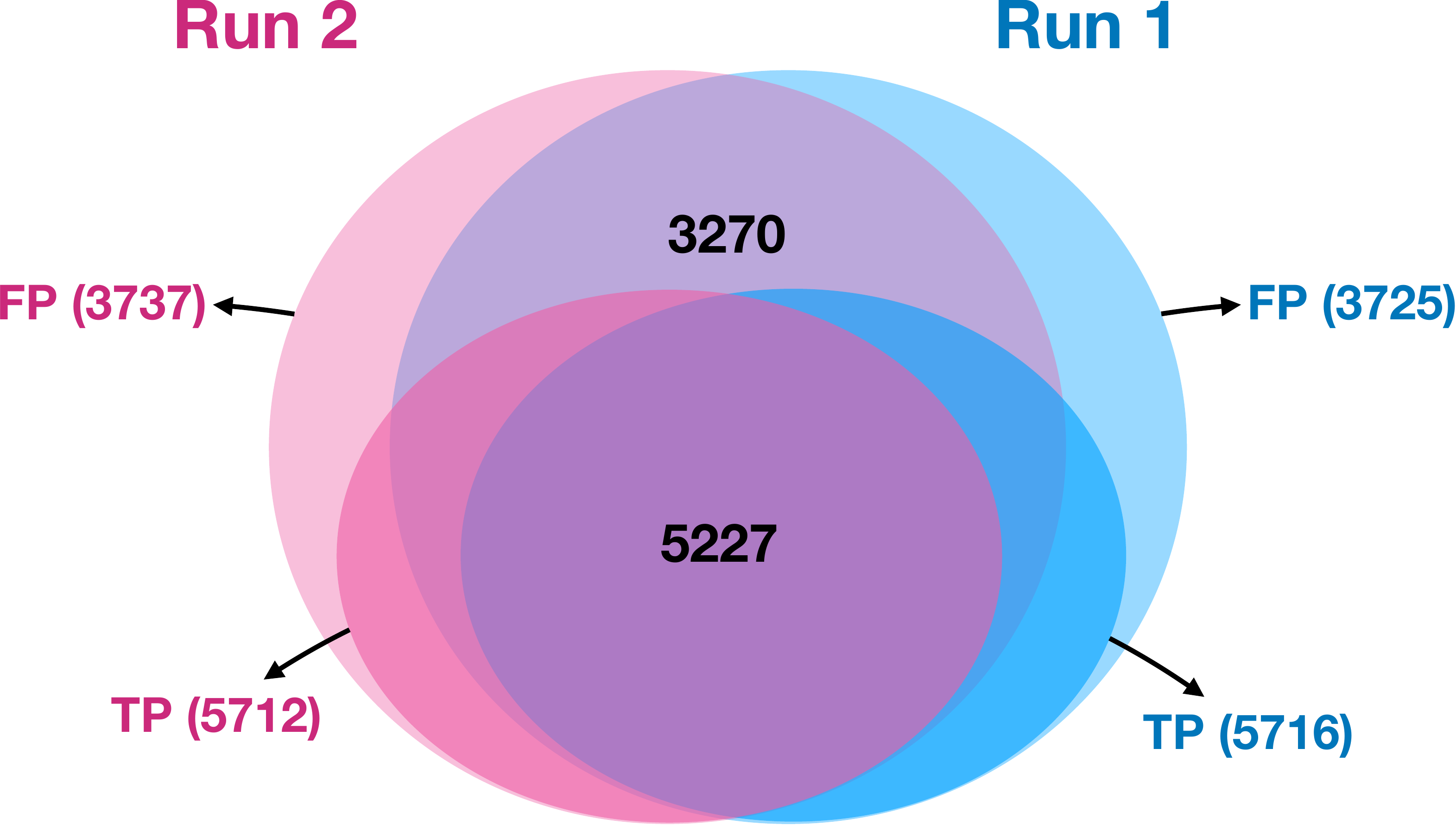}
    \caption{Overlap of membership predictions across two runs of logistic regression with RDP at $\epsilon = 1000$ (\dataset{Purchase-100})}
    \label{fig:purchase_lr_overlap}
\end{figure}

\begin{figure}[ptb]
    \centering
    \includegraphics[width=\linewidth]{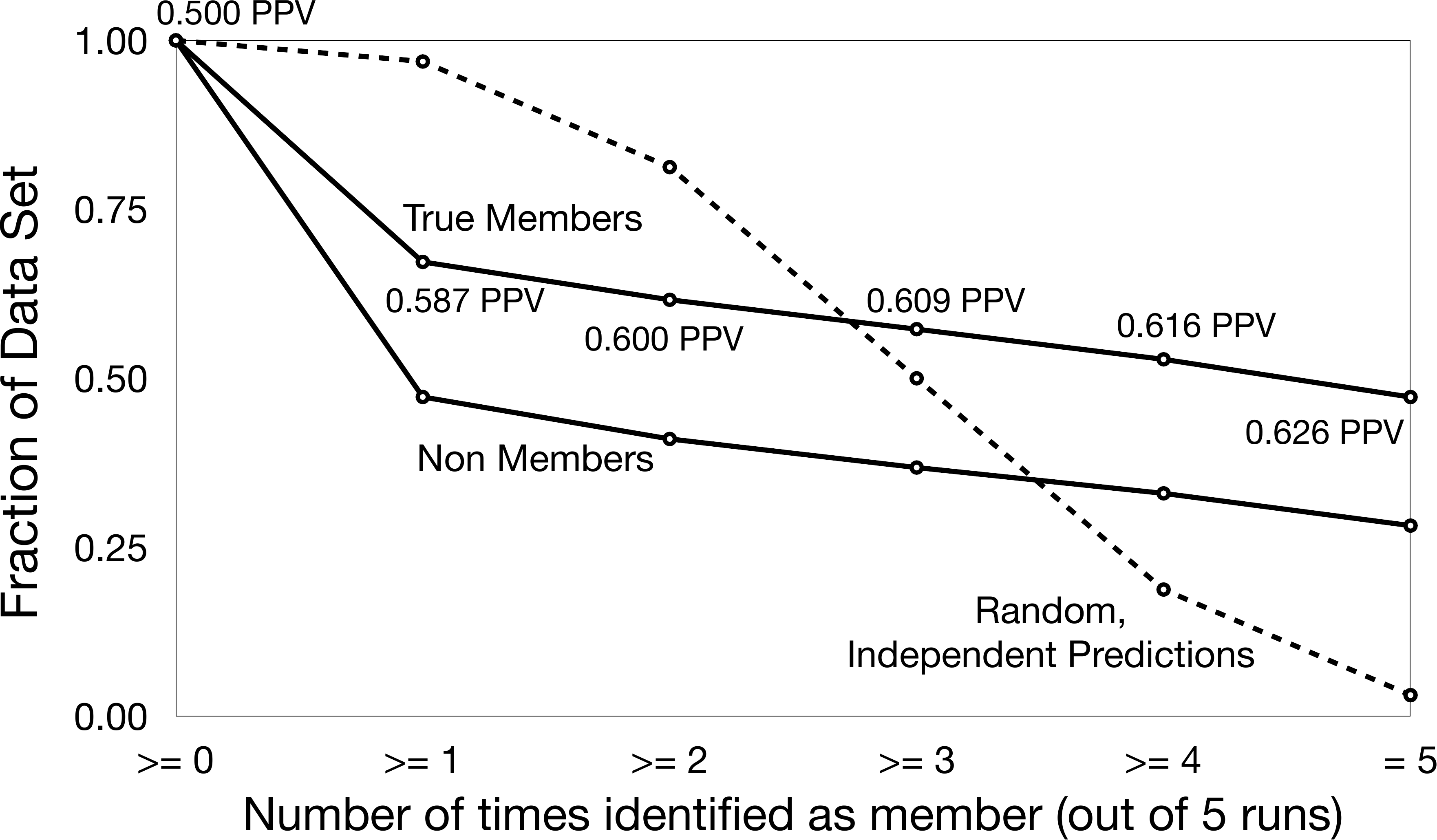}
    \caption{Membership predictions across multiple runs of logistic regression with RDP at $\epsilon = 1000$ (\dataset{Purchase-100})}
    \label{fig:purchase_lr_multiple_runs}
\end{figure}

\shortsection{\bfdataset{Purchase-100}}
For the \dataset{Purchase-100} dataset, the baseline model for non-private logistic regression achieves accuracy of 0.942 on the training set and 0.695 on test set. In comparison, Google ML platform's black-box trained model achieves a test accuracy of 0.656 (see Shokri et al.~\cite{shokri2017membership} for details).

Figure~\ref{fig:purchase_lr_grad_acc} shows the accuracy loss for all the differential privacy variants on \dataset{Purchase-100}. Na\"{i}ve composition and advanced composition have essentially no utility until $\epsilon$ exceeds 100. At $\epsilon = 1000$, na\"{i}ve composition achieves accuracy loss of $0.116 \pm 0.003$, advanced composition achieves accuracy loss of $0.513 \pm 0.003$, and the other variants achieve accuracy loss close to 0.02. As expected, RDP achieves the best utility across all $\epsilon$ values. 

Figure~\ref{fig:purchase_lr_grad_inference} compares the privacy leakage of the variants against the inference attacks. The leakage is in accordance to the noise each variant adds and increases proportionally with model utility. Hence, if a model has reasonable utility, it is bound to leak some membership information. The membership inference attack of Yeom et al.\ is relatively more effective than the Shokri et al.\ attack.

\begin{figure*}[ptb]
    \centering
    \subfigure[\dataset{CIFAR-100}]{
    \includegraphics[width=.45\textwidth]{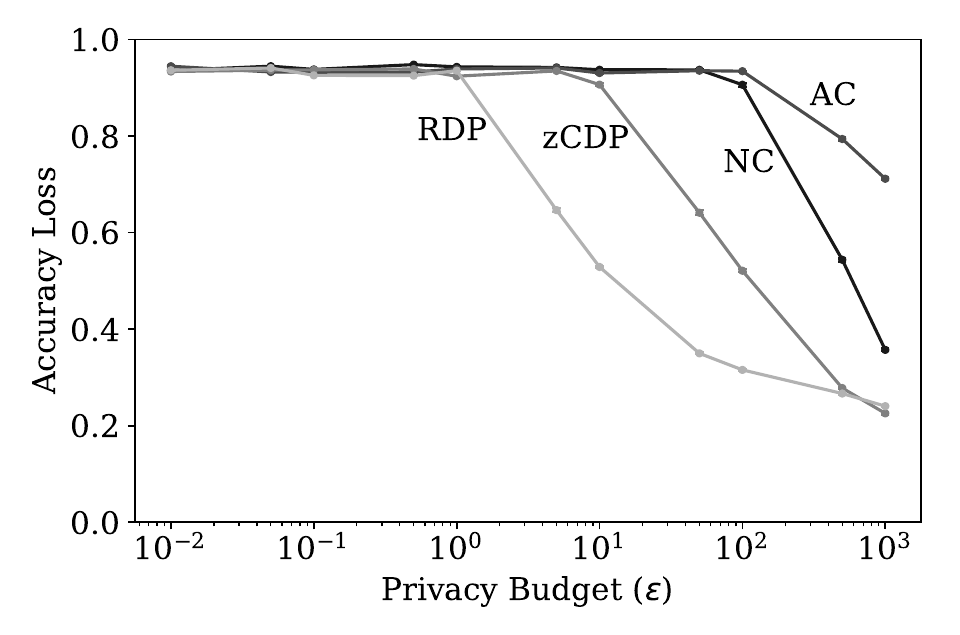}
    \label{fig:cifar_nn_grad_acc}}
    \subfigure[\dataset{Purchase-100}]{
    \includegraphics[width=.45\textwidth]{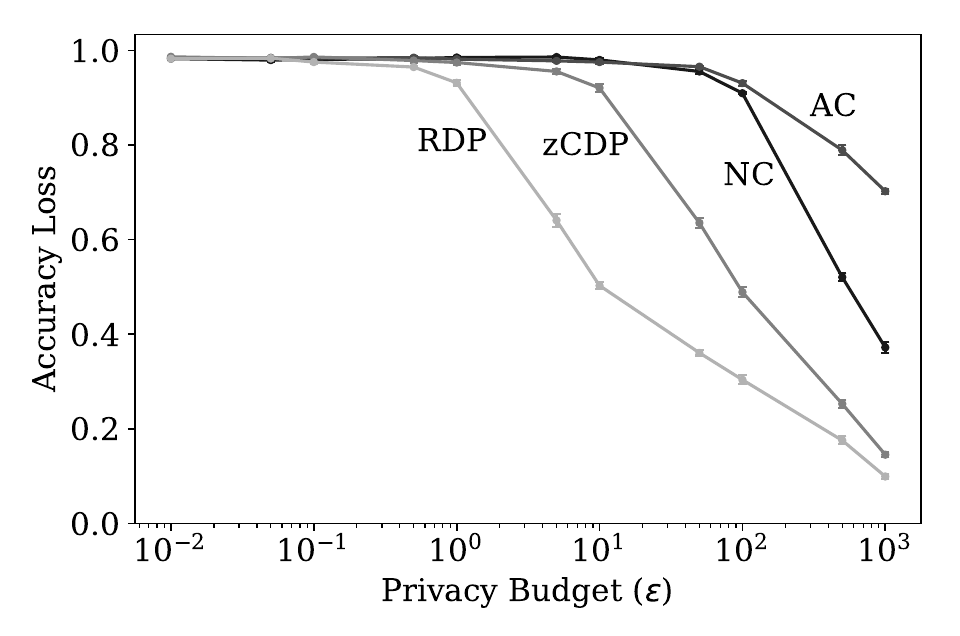}    \label{fig:purchase_nn_grad_acc}}
    \caption{Accuracy loss of neural networks.}
    \label{fig:both_nn_grad_acc}
\end{figure*}

\begin{figure*}[tbp]
\centering
    \subfigure[Shokri et al.\ membership inference]{
    \includegraphics[width=0.32\textwidth]{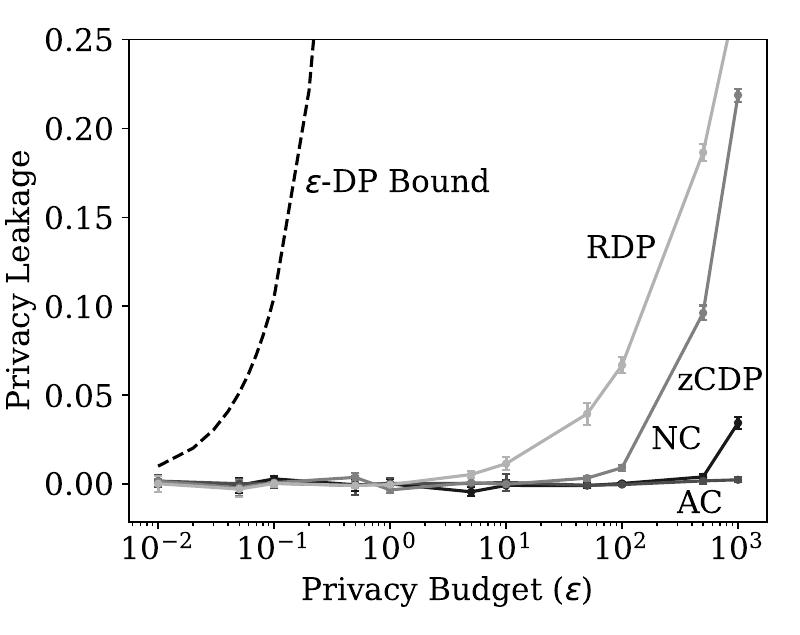}
    \label{fig:cifar_nn_grad_inference_attack}}
    \subfigure[Yeom et al.\ membership inference]{
    \includegraphics[width=0.32\textwidth]{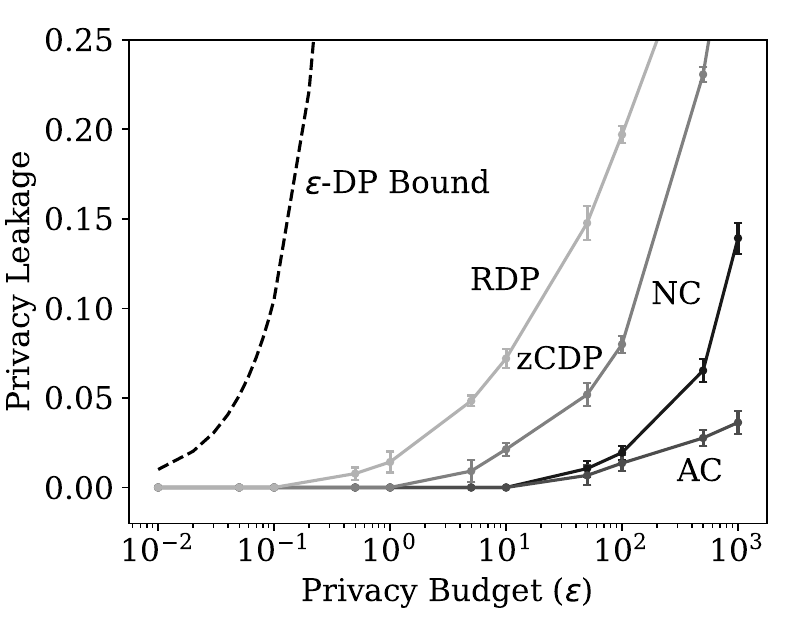}
    \label{fig:cifar_nn_grad_inference_mem}}
    \subfigure[Yeom et al.\ attribute inference]{
    \includegraphics[width=0.32\textwidth]{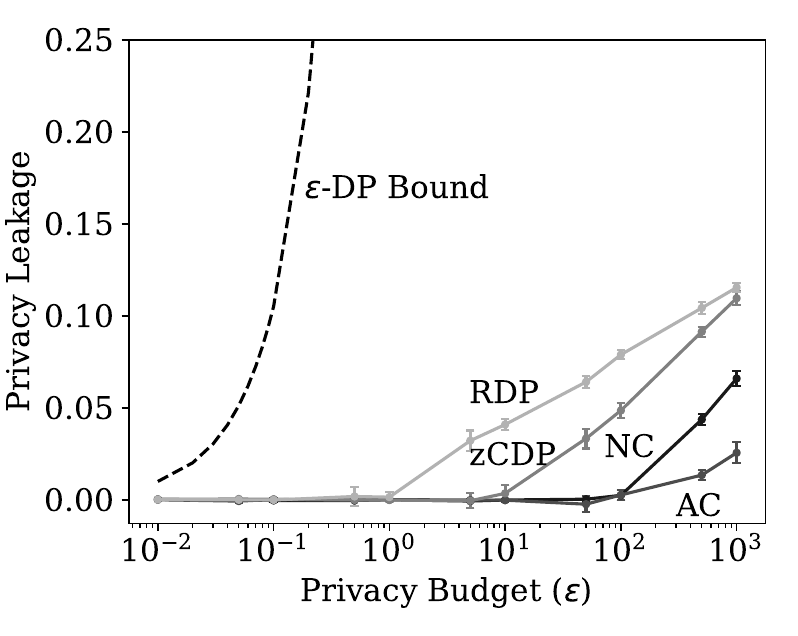}
    \label{fig:cifar_nn_grad_inference_attr}}
\caption{Inference attacks on neural network (\dataset{CIFAR-100}).}
\label{fig:cifar_nn_grad_inference}
\end{figure*}

\begin{figure*}[ptb]
\centering
\subfigure[Overlap of membership predictions across two runs]{
    \includegraphics[width=0.48\textwidth]{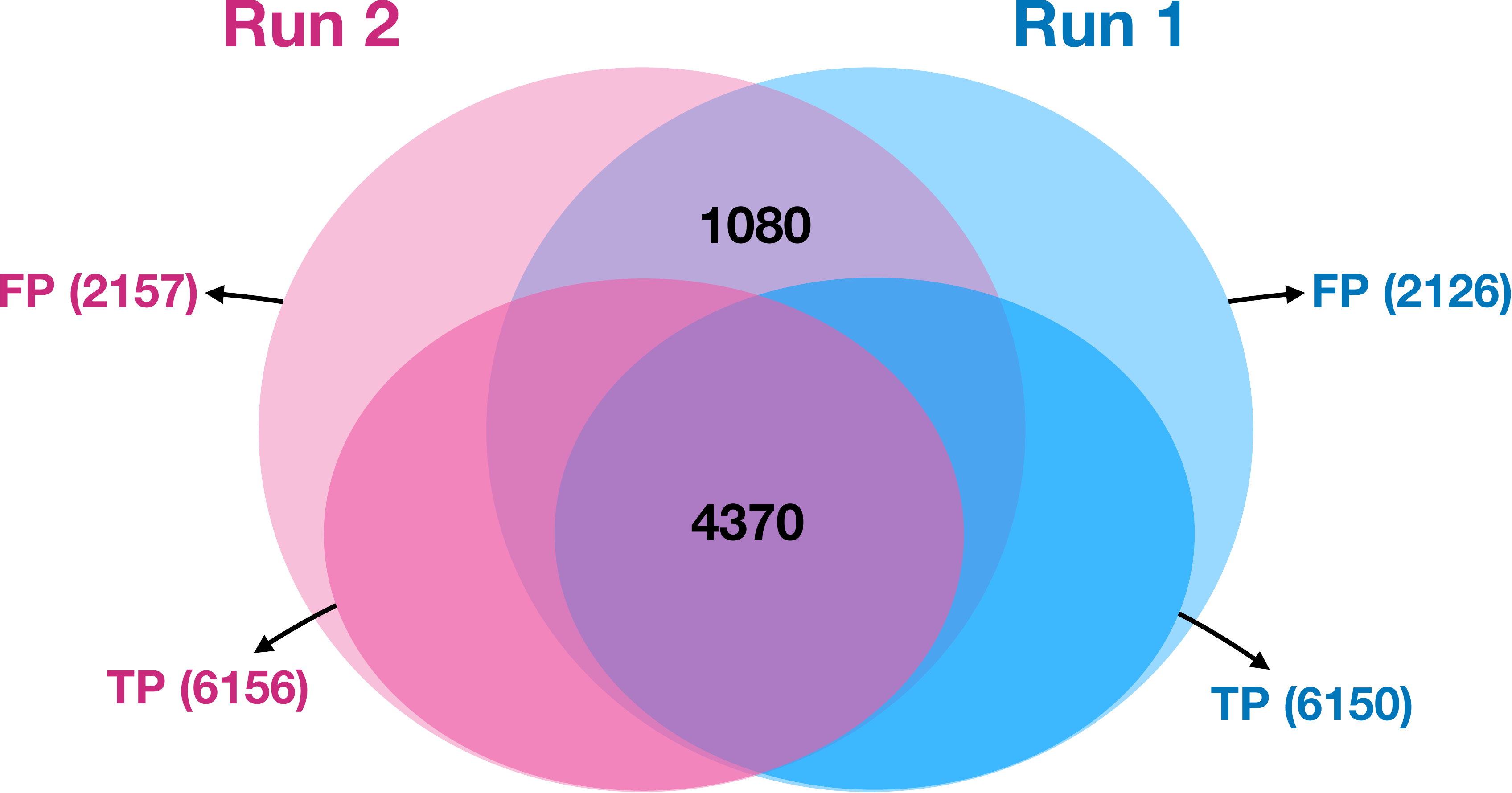}
    \label{fig:cifar_nn_overlap}}
\subfigure[Predictions across multiple runs]{
        \includegraphics[width=0.48\textwidth]{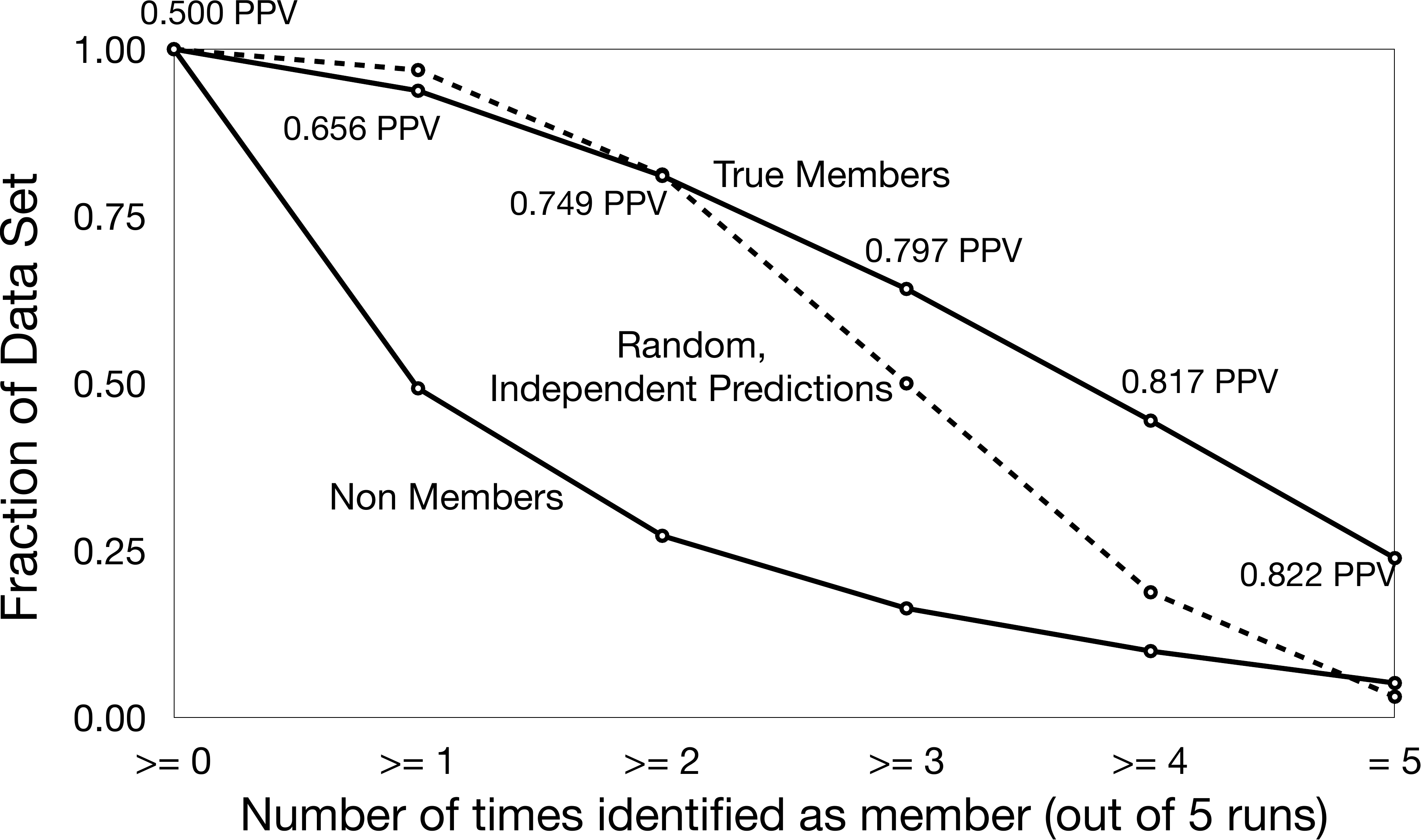}
        \label{fig:cifar_nn_multiple_runs}
    }
    \caption{Predictions across multiple runs of neural network with RDP at $\epsilon = 1000$ (\dataset{CIFAR-100})}\label{fig:cifar_nn_multiple_predcitions}
\end{figure*}

As with for \dataset{CIFAR-100} data set, we analyze the vulnerability of members across multiple runs of logistic regression model trained with RDP at $\epsilon = 1000$. Figure~\ref{fig:purchase_lr_overlap} shows the overlap of membership predictions across two runs, where the PPV increases from 0.605 for individual runs to 0.615 for the combination of runs. Similarly to \dataset{CIFAR-100}, the huge overlap of membership predictions indicates that the predictions are not random across individual runs but the small increase in PPV suggests that the properties that put records at risk of identification are not strongly correlated with being training set members. Figure~\ref{fig:purchase_lr_multiple_runs} shows the fraction of members and non-members that are repetitively predicted as members by the adversary across multiple runs. The adversary correctly predicts 4,723 members in all the five runs with a PPV of 0.63, meaning that these members are more vulnerable to the inference attack. In contrast, the adversary correctly identifies 6,008 members using the non-private model with a PPV of 0.62. In this setting, the privacy leakage is considerable and the PPV of 0.62 poses some threat in situations where adversaries may also have access to some additional information. 

\subsection{Neural Networks}\label{sec:results:nn}
We train neural network models consisting of two hidden layers and an output layer. The hidden layers have 256 neurons that use ReLU activation. The output layer is a softmax layer with 100 neurons, each corresponding to a class label. This architecture is similar to the one used by Shokri et al.~\cite{shokri2017membership}.

\shortsection{\bfdataset{CIFAR-100}}
The baseline non-private neural network model achieves accuracy of 1.000 on the training set and 0.168 on test set, which is competitive to the neural network model of Shokri et al.~\cite{shokri2017membership}. Their model is trained on a training set of size 29,540 and achieves test accuracy of 0.20, whereas our model is trained on 10,000 training instances. 
There is a huge generalization gap of 0.832, which the inference attacks can exploit. Figure~\ref{fig:cifar_nn_grad_acc} compares the accuracy loss of neural network models trained with different notions of differential privacy with varying privacy budget $\epsilon$. The model trained with na\"{i}ve composition does not learn anything useful until $\epsilon = 100$ (accuracy loss of $0.907 \pm 0.004$), at which point the advanced composition also has accuracy loss close to 0.935 and the other variants achieve accuracy loss close to 0.24. None of the variants approach zero accuracy loss, even for $\epsilon = 1000$. The relative performance is similar to that of the logistic regression model discussed in Section~\ref{sec:results:erm}.

\begin{figure*}[tb]
\centering
    \subfigure[Shokri et al.\ membership inference]{
    \includegraphics[width=0.32\textwidth]{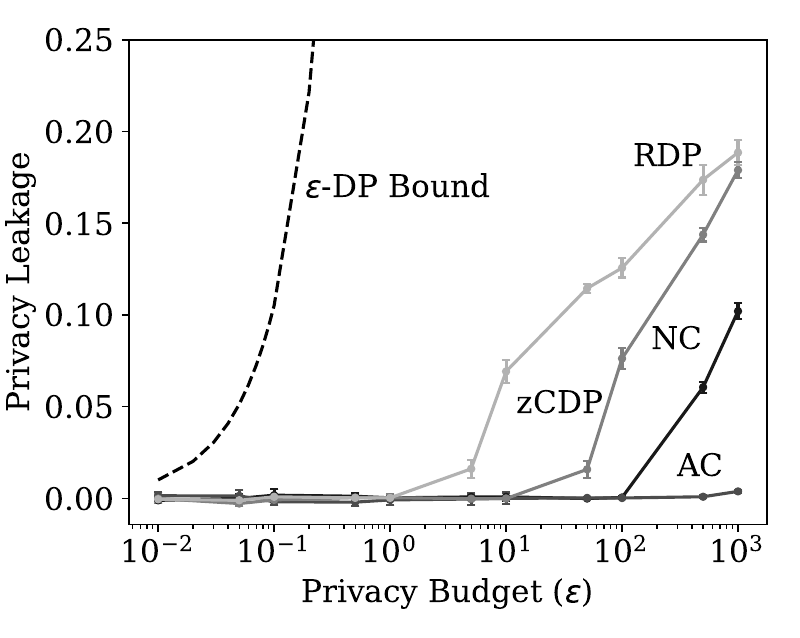}
    \label{fig:purchase_nn_grad_inference_attack}}
    \subfigure[Yeom et al.\ membership inference]{
    \includegraphics[width=0.32\textwidth]{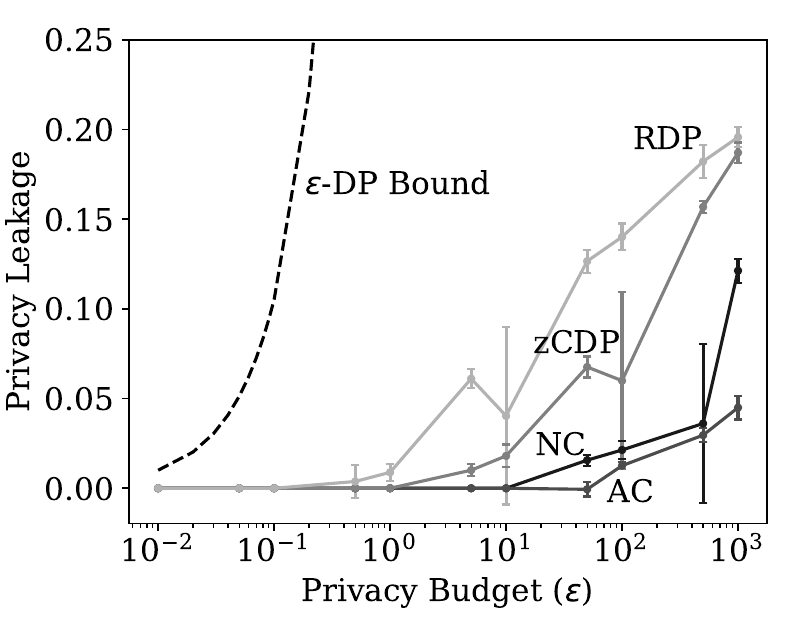}
    \label{fig:purchase_nn_grad_inference_mem}}
    \subfigure[Yeom et al.\ attribute inference]{
    \includegraphics[width=0.32\textwidth]{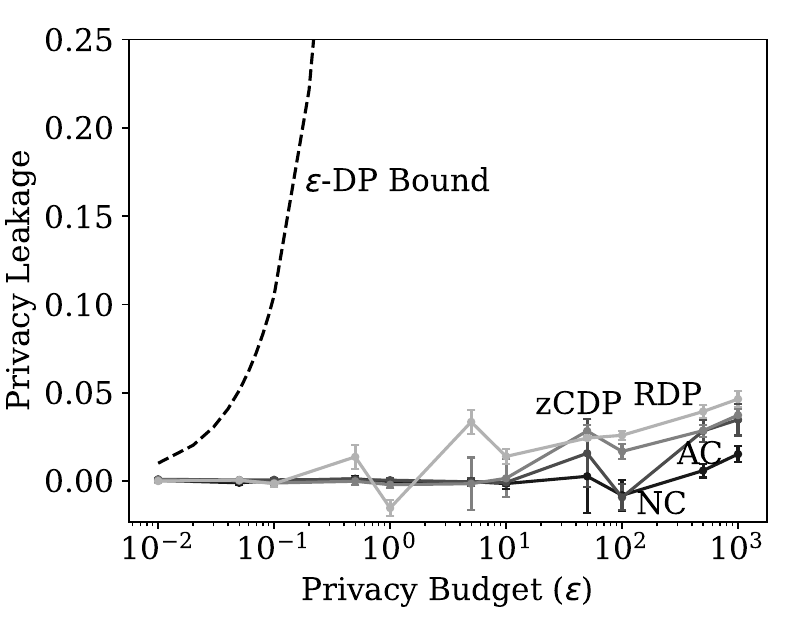}
    \label{fig:purchase_nn_grad_inference_attr}}
\caption{Inference attacks on neural network (\dataset{Purchase-100}).}
\label{fig:purchase_nn_grad_inference}
\end{figure*}

\begin{figure*}[ptb]
\centering
\subfigure[Overlap of membership predictions across two runs]{
    \includegraphics[width=0.48\textwidth]{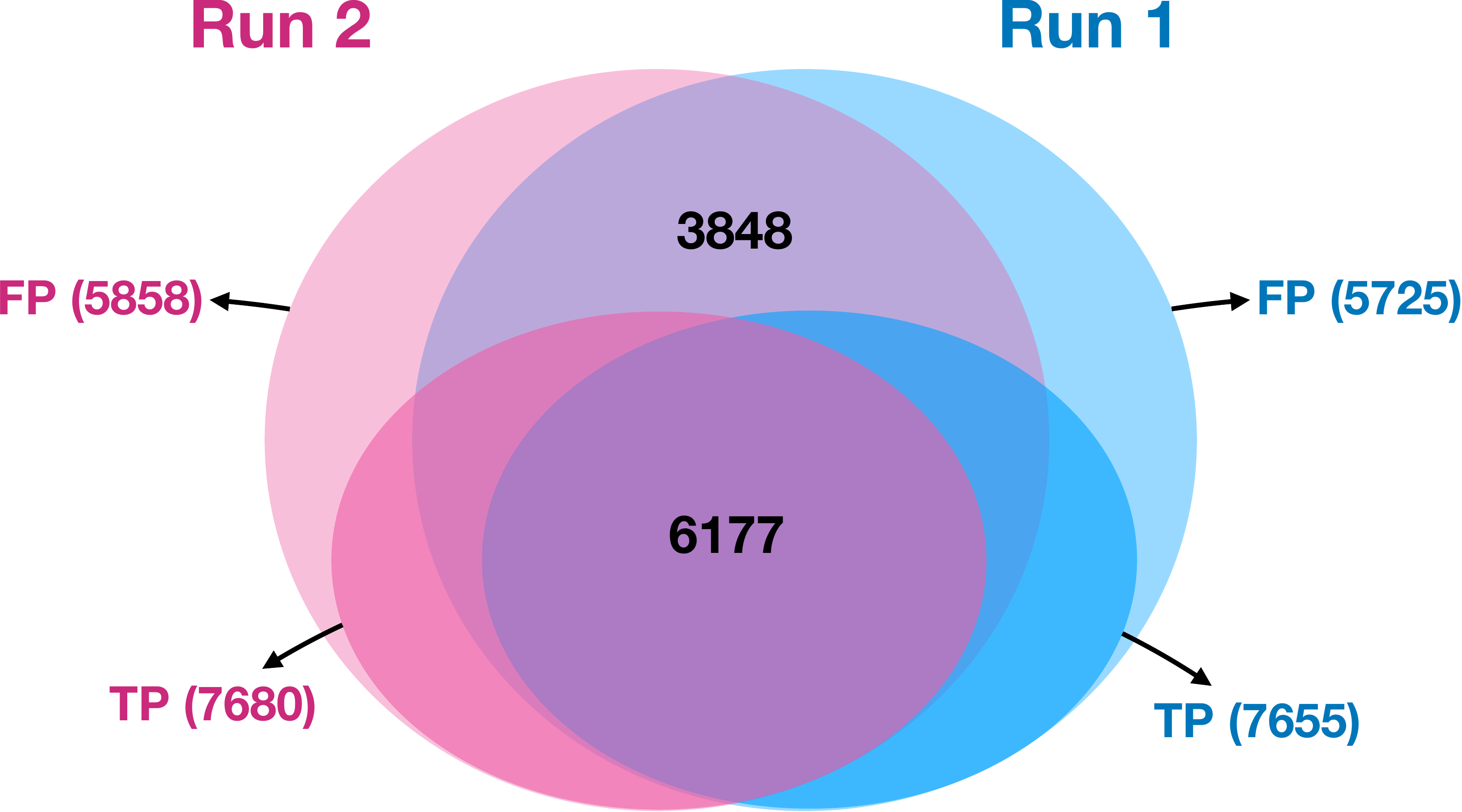}
    \label{fig:purchase_nn_overlap}}
\subfigure[Predictions across multiple runs]{
\   \includegraphics[width=0.48\textwidth]{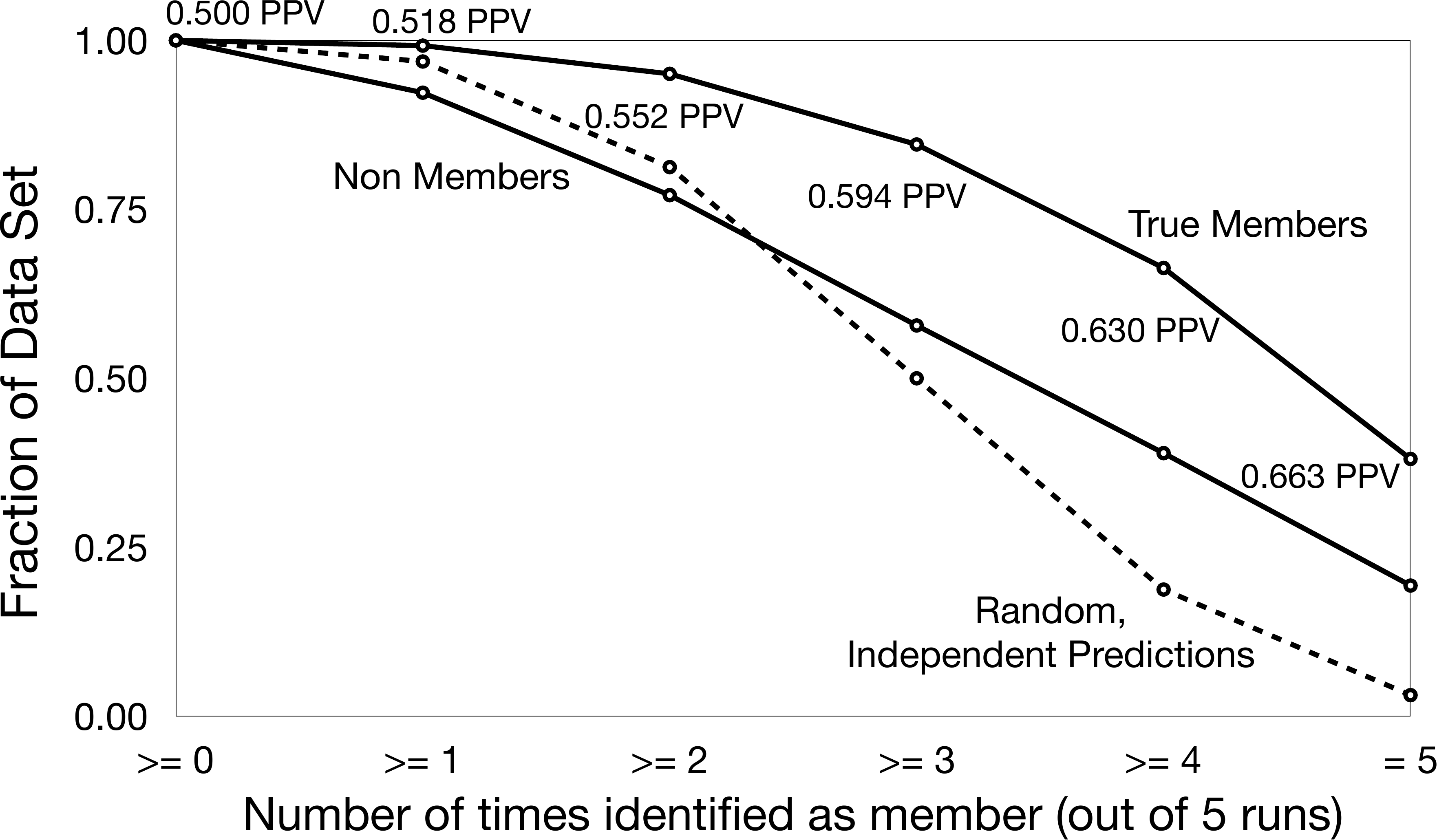}
        \label{fig:purchase_nn_multiple_runs}
    }
    \caption{Predictions across multiple runs of neural network with RDP at $\epsilon = 1000$ (\dataset{Purchase-100})}
\end{figure*}

Figure~\ref{fig:cifar_nn_grad_inference} shows the privacy leakage on neural network models for the inference attacks. The privacy leakage for each variant of differential privacy accords with the amount of noise it adds to the model. At higher $\epsilon$ values, the leakage is significant for zCDP and RDP due to model overfitting. For $\epsilon = 1000$, with the Shokri et al.\ attack, na\"{i}ve composition has leakage of $0.034$ compared to 
$0.219$ for zCDP and $0.277$ for RDP (above the region shown in the plot). For the Yeom et al.\ attack, at $\epsilon = 1000$, RDP exhibits privacy leakage of $0.399$, a substantial advantage for the adversary. Na\"{i}ve composition and advanced composition achieve strong privacy against membership inference attackers, but fail to learning anything useful even with the highest privacy budgets tested. RDP and zCDP are able to learn useful models at privacy budgets above $\epsilon = 100$, but exhibit significant privacy leakage at the corresponding levels. Thus, none of the \dataset{CIFAR-100} NN models seem to provide both acceptable model utility and strong privacy.

Like we did for logistic regression, we also analyze the vulnerability of individual records across multiple model training runs with RDP at $\epsilon = 1000$. Figure~\ref{fig:cifar_nn_multiple_predcitions} shows the membership predictions across multiple runs. In contrast to the logistic regression results, for the neural networks the fraction of repetitively exposed members decreases with increasing number of runs while the number of repeated false positives drops more drastically, leading to a significant increase in PPV. The adversary identifies 2,386 true members across all the five runs with a PPV of 0.82. In comparison, the adversary exposes 7,716 members with 0.94 PPV using the single non-private model. So, although the private models do leak significant information, the privacy noise substantially diminishes the inference attacks.

\shortsection{\bfdataset{Purchase-100}}
The baseline non-private neural network mod\-el achieves accuracy of 0.982 on the training set and 0.605 on test set. In comparison, the neural network model of Shokri et al.~\cite{shokri2017membership} trained on a similar data set (but with 600 attributes instead of 100 as in our data set) achieves 0.670 test accuracy.  Figure~\ref{fig:purchase_nn_grad_acc} compares the accuracy loss of neural network models trained with different variants of differential privacy.  The trends are similar to those for the logistic regression models  (Figure~\ref{fig:purchase_lr_grad_acc}). The zCDP and RDP variants achieve model utility close to the non-private baseline for $\epsilon = 1000$, while na\"{i}ve composition continues to suffer from high accuracy loss ($0.372$). Advanced composition has higher accuracy loss of $0.702$ for $\epsilon = 1000$ as it requires addition of more noise than na\"{i}ve composition when $\epsilon$ is greater than the number of training epochs.

Figure~\ref{fig:purchase_nn_grad_inference} shows the privacy leakage comparison of the variants against the inference attacks. The results are consistent with those observed for \dataset{CIFAR-100}. Similar to the models in other settings, we analyze the vulnerability of members across multiple runs of neural network model trained with RDP at $\epsilon = 1000$. Figure~\ref{fig:purchase_nn_overlap} shows the overlap of membership predictions across two runs, while the figure~\ref{fig:purchase_nn_multiple_runs} shows the fraction of members and non-members classified as members by the adversary across multiple runs. The PPV increases with intersection of more runs, but less information is leaked than for the \dataset{CIFAR-100} models. The adversary correctly identifies 3,807 members across all five runs with a PPV of 0.66. In contrast, the adversary predicts 9,461 members with 0.64 PPV using the non-private model.

\subsection{Discussion}
While the tighter cumulative privacy loss bounds provided by variants of differential privacy improve model utility for a given privacy budget, the reduction in noise increases vulnerability to inference attacks. While these definitions still satisfy the $(\epsilon,\delta)$-differential privacy guarantees, the meaningful value of these guarantees diminishes rapidly with high $\epsilon$ values.  Although the theoretical guarantees provided by differential privacy are very appealing, once $\epsilon$ values exceed small values, the practical value of these guarantees is insignificant---in most of our inference attack figures, the theoretical bound given by $\epsilon$-DP falls off the graph before any measurable privacy leakage occurs (and at levels well before models provide acceptable utility). The value of these privacy mechanisms comes not from the theoretical guarantees, but from the impact of the mechanism on what realistic adversaries can infer. Thus, for the same privacy budget, differential privacy techniques that produce tighter bounds and result in lower noise requirements come with increased concrete privacy risks. 

We note that in our inference attack experiments, we use equal numbers of member and non-member records which provides 50-50 prior success probability to the attacker. Thus, even an $\epsilon$-DP implementation might leak even for small $\epsilon$ values, though we did not observe any such leakage. Alternatively, a skewed prior probability may lead to smaller leakage even for large $\epsilon$ values. Our goal in this work is to evaluate scenarios where risk of inference is high, so the use of 50-50 prior probability is justified. We also emphasize that our results show the privacy leakage due to three particular inference attacks. Attacks only get better, so future attacks may be able to infer more than is shown in our experiments. 

\section{Conclusion}

Differential privacy has earned a well-deserved reputation providing principled and powerful mechanisms for establishing privacy guarantees. However, when it is implemented for challenging tasks such as machine learning, compromises must be made to preserve utility. It is essential that the privacy impact of those compromises is well understood when differential privacy is deployed to protect sensitive data. Our results reveal that the commonly-used combinations of $\epsilon$ values and the variations of differential privacy in practical implementations may provide unacceptable utility-privacy trade-offs.  We hope our study will encourage more careful assessments of the practical privacy value of formal claims based on differential privacy, and lead to deeper understanding of the privacy impact of design decisions when deploying differential privacy. There remains a huge gap between what the state-of-the-art inference attacks can infer, and the guarantees provided by differential privacy. Research is needed to understand the limitations of inference attacks, and eventually to develop solutions that provide desirable, and well understood, utility-privacy trade-offs.

\section*{Availability} 
Open source code for reproducing all of our experiments is available at \url{https://github.com/bargavj/EvaluatingDPML}.

\section*{Acknowledgments} 
The authors are deeply grateful to {\'U}lfar Erlingsson for pointing out some key misunderstandings in an early version of this work and for convincing us of the importance of per-instance gradient clipping, and to {\'U}lfar, Ilya Mironov, and Shuang Song for help validating and improving the work. We thank Brendan McMahan for giving valuable feedback and important suggestions on improving the clarity of the paper. We thank Vincent Bindschaedler for shepherding our paper. We thank Youssef Errami and Jonah Weissman for contributions to the experiments, and Ben Livshits for feedback on the work. Atallah Hezbor, Faysal Shezan, Tanmoy Sen, Max Naylor, Joshua Holtzman and Nan Yang helped systematize the related works. Finally, we thank Congzheng Song and Samuel Yeom for providing their implementation of inference attacks. This work was partially funded by grants from the National Science Foundation SaTC program (\#1717950, \#1915813) and support from Intel and Amazon.

\bibliographystyle{plain}
\bibliography{refusenix}

\end{document}